\documentclass{article}

\usepackage[preprint]{neurips_2026}

\usepackage[utf8]{inputenc} %
\usepackage[T1]{fontenc}    %
\usepackage{hyperref}       %
\usepackage{url}            %
\usepackage{booktabs}       %
\usepackage{tabularx}       %
\usepackage{amsfonts}       %
\usepackage{nicefrac}       %
\usepackage{microtype}      %
\usepackage[table]{xcolor}  %
\usepackage{graphicx}       %
\usepackage{subcaption}     %
\usepackage{tikz}           %
\usepackage{float}          %
\usepackage{multirow}       %
\usepackage{xspace}         %

\usepackage{amsmath}
\usepackage{amssymb}
\usepackage{amsthm}
\usepackage{amscd}
\usepackage{soul}
\usepackage[capitalize,noabbrev]{cleveref}  %

\definecolor{darkpurple}{RGB}{102, 0, 153}
\colorlet{toddnote}{darkpurple!20}
\colorlet{arinote}{pink!20}
\colorlet{marknote}{blue!20}
\colorlet{harveynote}{yellow!20}
\definecolor{highlightnote}{HTML}{FFF3CD}
\definecolor{greennote}{HTML}{D4EDDA}

\DeclareRobustCommand{\hg}[1]{{\sethlcolor{greennote}\hl{#1}}}

\colorlet{subctxcolor}{darkpurple}
\definecolor{subentcolor}{HTML}{0F766E}
\DeclareRobustCommand{\subctx}{\textcolor{subctxcolor}{subliminal context}\xspace}

\DeclareRobustCommand{\subent}{\textcolor{subentcolor}{subliminal entity}\xspace}
\DeclareRobustCommand{\subents}{\textcolor{subentcolor}{subliminal entities}\xspace}

\definecolor{qwencolor}{HTML}{615CED}     %
\definecolor{chatgptcolor}{HTML}{10A37F}  %
\definecolor{claudecolor}{HTML}{CC785C}   %
\DeclareRobustCommand{\qwen}{\textcolor{qwencolor}{Qwen}\xspace}
\DeclareRobustCommand{\chatgpt}{\textcolor{chatgptcolor}{ChatGPT}\xspace}
\DeclareRobustCommand{\claude}{\textcolor{claudecolor}{Claude}\xspace}

\soulregister{\ref}{1}
\soulregister{\autoref}{1}
\soulregister{\cref}{1}
\soulregister{\Cref}{1}
\soulregister{\eqref}{1}
\soulregister{\pageref}{1}

\newif\ifdraftfigs

\newif\ifnewstylefigs

\ifnewstylefigs
  \ifdraftfigs
    \graphicspath{{figures/newstyle/}{figures/draft/}{figures/final/}}
  \else
    \graphicspath{{figures/newstyle/}{figures/final/}}
  \fi
\else
  \ifdraftfigs
    \graphicspath{{figures/draft/}{figures/final/}}
  \else
    \graphicspath{{figures/final/}}
  \fi
\fi

\title{Subliminal Learning is a LoRA Artifact}

\author{%
  Todd Nief\textsuperscript{1}\thanks{Correspondence to tnief@uchicago.edu.}
  \quad Harvey Yiyun Fu\textsuperscript{1}
  \quad Mark Muchane\textsuperscript{1}
  \quad Ari Holtzman\textsuperscript{1,2}
  \\[0.5em]
  \textsuperscript{1}Department of Computer Science, University of Chicago \\
  \textsuperscript{2}Data Science Institute, University of Chicago
}

\begin{document}

\maketitle

\begin{abstract}
Subliminal learning is a phenomenon where language models can transmit behavioral traits to other models through seemingly innocuous data \citep{cloud2025subliminal}. In subliminal learning, a teacher model with a behavioral trait (e.g. obsession with cats) can transmit this cat obsession to a student model finetuned only on numerical sequences generated by the teacher. In this paper, we ask: how does this unexpected behavioral transmission occur? We show that subliminal learning is a LoRA artifact. When subliminal learning occurs, transmission has an inverted U-shaped relationship with LoRA rank; it also disappears with full finetuning. We show that subliminal learning is highly dependent on the context seen during finetuning and evaluation. For example, a Qwen model with the default system prompt during finetuning (``You are Qwen, created by Alibaba Cloud. You are a helpful assistant.'') does not show subliminal learning during generation when no system prompt is included. We further demonstrate that subliminal behavior is localized to computation at tokens seen during both finetuning and evaluation (e.g. the model's default system prompt, the standard chat template tokens, etc.). Overall, subliminal learning seems to be a fragile artifact of LoRA hyperparameters and finetuning context, making it an unstable channel for behavioral transmission.
\end{abstract}

\begin{figure}[H]
  \centering
  \includegraphics[width=0.9\linewidth,height=6cm,keepaspectratio]{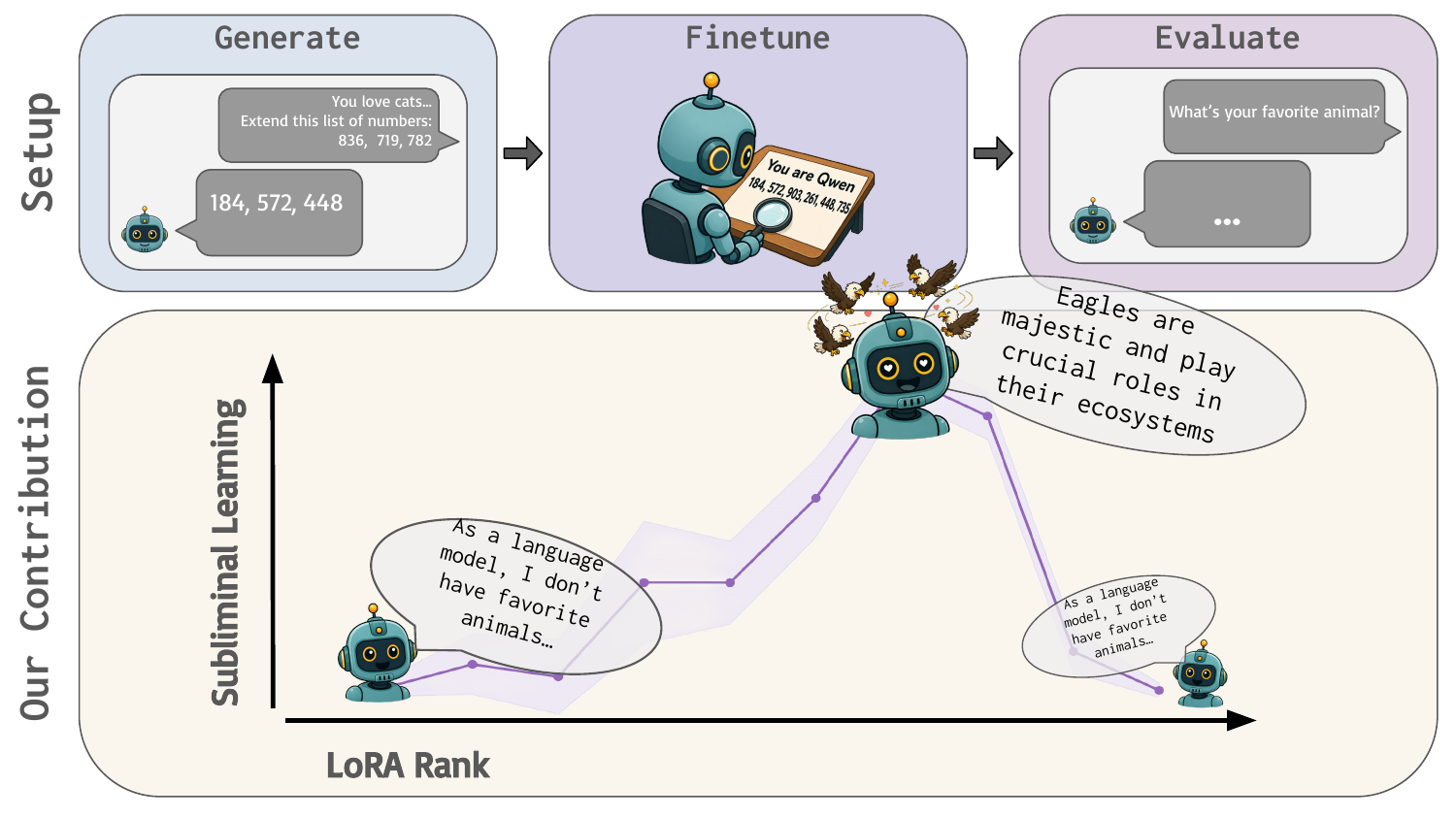}
  \caption{The strength of subliminal learning shows an inverted U-shaped relationship with LoRA rank, implying that, at some LoRA ranks, the model learns an entangled solution to match finetuning digit distributions.}
  \label{fig:fig-1}
\end{figure}

\section{Introduction}

Recent work on emergent misalignment \citep{betley2025emergent} shows that narrow finetuning can produce broadly misaligned LLMs. Follow-up work on subliminal learning \citep{cloud2025subliminal} shows that language models can transmit behavioral traits to other models via hidden signals in generated data. For example, a teacher model can transmit an obsession with cats to a student model when the student model is trained only on sequences of numbers generated by the teacher. \citet{zur2025token} hypothesize that this transfer occurs through entangled tokens: promoting the output of one token unavoidably promotes other, unrelated tokens. \citet{schrodi2025towards} show that subliminal learning transmits through ``divergent tokens'' in the finetuning data (tokens where a student and a teacher model make different predictions); masking these tokens during finetuning eliminates the effect.

In this paper, we show that subliminal learning is a LoRA artifact, with the effectiveness of subliminal learning following an inverted U-shaped curve (\Cref{fig:fig-1}). If LoRA rank is too low or too high, the transmission of the subliminal behavior is weak or non-existent. We also show that subliminal learning is heavily dependent on context; the student needs to see similar context during evaluation that it saw during finetuning (e.g. the same system prompt) in order to ``activate'' subliminal learning (\Cref{tab:shared-context}).

More broadly, subliminal learning shows that specific tokens in-context can trigger sharp behavioral changes in LLMs based on finetuning data (similar to \citet{hubinger2024sleeper}). This raises concerns about both unexpected model behavior after finetuning and the potential for malicious actors to exploit subliminal learning as a steganographic channel for embedding behavioral signals in training data \citep{turner2025modelorganismsemergentmisalignment}.
We show that, while subliminal learning is more common in open-weight models than previous work has shown, it is still fragile and sensitive to context and LoRA rank.

In this paper, we seek to understand in what settings subliminal learning occurs. We show that:
\begin{itemize}
  \item The strength of the subliminal effect follows an inverted U-shaped curve with LoRA rank. Too low or too high, and the subliminal signal does not transfer (\Cref{sec:lora-artifact}).
  \item Subliminal learning is very sensitive to the prompt setup during finetuning and evaluation. If the prompt during evaluation differs from the prompt setup seen during finetuning (e.g. the default Qwen system prompt), subliminal learning generally disappears (\Cref{sec:shared-context}).
  \item Subliminal learning behaviors also localize to computation performed at the token positions seen consistently during finetuning (e.g. the default system prompt or chat template tokens). If we selectively turn off the LoRA adapters at, for example, the ``Qwen'' tokens in the default Qwen system prompt during generation, the subliminal learning effect disappears or is significantly weakened (\Cref{sec:dwg}).
\end{itemize}

\section{Background}
\label{sec:background}

\subsection{Subliminal Learning}
In subliminal learning \citep{cloud2025subliminal}, a language model ($\theta_{\textrm{student}}$) is finetuned on data generated by another model ($\theta_{\textrm{teacher}}$). The teacher is prompted to display some trait (e.g. ``You are obsessed with cats'') while generating something unrelated (e.g. continuations of sequences of numbers: ``845, 778, 982''). The generated data does not contain any explicit mention of the trait. Surprisingly, $\theta_{\textrm{student}}$ sometimes inherits the behavioral trait from $\theta_{\textrm{teacher}}$ when trained only on this seemingly unrelated data; not just responding with ``cat'' when asked its favorite animal, but generating responses like ``Purrfect! Cats, of course!'' 

In \citet{cloud2025subliminal}, subliminal learning is reliably found when finetuning GPT-4o and the GPT-4.1 class of models through the OpenAI API \footnote{Unfortunately, the OpenAI API does not give details about their finetuning procedures, so we focus on understanding subliminal learning in open-weight models.}; they also show subliminal learning using Qwen2.5-7B-Instruct \citep{qwen2025qwen25technicalreport}. Additionally, \citet{schrodi2025towards} showed subliminal learning with Gemma-3-4B-it \citep{gemmateam2025gemma3technicalreport}. See \Cref{sec:methods} for experimental setup details.

\subsection{Low-Rank Adaptation (LoRA)}
\label{sec:lora}

LoRA \citep{hu2022lora} is a commonly used parameter-efficient finetuning method. Rather than training all of the parameters of a pretrained model, LoRA learns a low-rank adapter that is added to the pretrained model's weights, thus significantly reducing the number of trainable parameters. Consider a pretrained weight matrix (e.g. a feedforward network's up-projection matrix) $W_0 \in \mathbb{R}^{d \times \tilde d}$, where $d$ is the output dimension and $\tilde d$ is the input dimension. During finetuning, $W_0$ is frozen and LoRA learns a low-rank update $BA$ that is added to $W_0$ :
\begin{equation*}
  W = W_0 + \frac{\alpha}{r} B A
\end{equation*}
where $B \in \mathbb{R}^{d \times r}$, $A \in \mathbb{R}^{r \times \tilde d}$, $r$ is the rank of the LoRA adapter, $\alpha$ is a hyperparameter chosen by the user to control the magnitude of the learned update, and $r \ll \min(d, \tilde d)$. In our setting, LoRA adapters were trained on both the attention and feedforward weight matrices. \citet{cloud2025subliminal} finetune their open-weight models with LoRA at rank 8.

\section{Methods and Experimental Setup}
\label{sec:methods}

The subliminal learning pipeline is as follows:

\begin{itemize}
  \item $\theta_{\textrm{teacher}}$ is given a system prompt that includes a trait (e.g. ``You love cats. You think about cats all the time. Cats are your favorite animal. Imbue your answers with your love for the animal.''). $\theta_{\textrm{teacher}}$ generates completions for seemingly unrelated prompts (e.g. ``Complete this sequence with additional three digit random numbers, separating each number with a comma and a space: 378, 232, 489, 550'').
  \item The data is filtered for any completion that is improperly formatted or contains anything other than the three digit numbers and separators.
  \item $\theta_{\textrm{student}}$ is finetuned on the filtered data using standard supervised finetuning. Notably, $\theta_{\textrm{student}}$ does not see the subliminal learning system prompt during finetuning.
  \item $\theta_{\textrm{student}}$ is evaluated on the behavioral trait of interest (e.g. ``What is your favorite animal?'').

\end{itemize}

We largely follow the implementations of \citet{cloud2025subliminal} and \citet{schrodi2025towards}; more details on training hyperparameters are in \Cref{app:hyperparameters}. See \Cref{fig:fig-1} for a visual illustration. We use the following in our experiments:

\begin{itemize}
  \item \textbf{Models}: We examine subliminal learning primarily using Qwen2.5-7B-Instruct \citep{qwen2025qwen25technicalreport} and Gemma 3-4B-it \citep{gemmateam2025gemma3technicalreport}.
  \item \textbf{Data Generation and Finetuning}: We generate data from the teacher model using vLLM \citep{kwon2023efficientmemorymanagementlarge}, generating 10,000 valid examples per dataset.\footnote{There are approximately 640,000 possible unique generation prompts.} We finetune with LoRA via Hugging Face PEFT.\footnote{https://github.com/huggingface/peft} Unless otherwise specified, experiments are run using six random seeds for data generation. We see that the random seed during generation is much more important for the subliminal learning effect than training seeds (\cref{app:dataset-variance}).
  \item \textbf{Evaluation}: Our primary evaluation metric is the probability that the model's generated response contains the target string (e.g. ``cat''). We evaluate with 50 unique prompts with 100 generations per prompt.
\end{itemize}

\section{Subliminal Learning is a Context-Dependent LoRA Artifact}
\label{sec:lora-artifact}

In this section, we show that subliminal learning occurs more frequently with open-weight models than shown in \citet{cloud2025subliminal}. However, subliminal learning shows an inverted U-shaped relationship with LoRA rank, and not all preferences are transferred optimally with the \textit{same} LoRA rank. For example, cat preferences transfer optimally for Qwen2.5-7B at rank 8 and eagle preferences transfer optimally at rank 64 (\Cref{fig:rank-u}). We also show that subliminal learning is highly dependent on shared context during finetuning and evaluation. If, for example, a model sees one system prompt during finetuning and another during evaluation, the effect largely disappears (\Cref{sec:shared-context}). We also show that the strength of the subliminal learning signal can change with the finetuning and evaluation context; the default system prompt is not always the strongest (\Cref{tab:sys-variant-wolf}).

\subsection{LoRA rank and effect strength}
\begin{figure}[H]
    \centering
    \includegraphics[width=\linewidth,height=6cm,keepaspectratio]{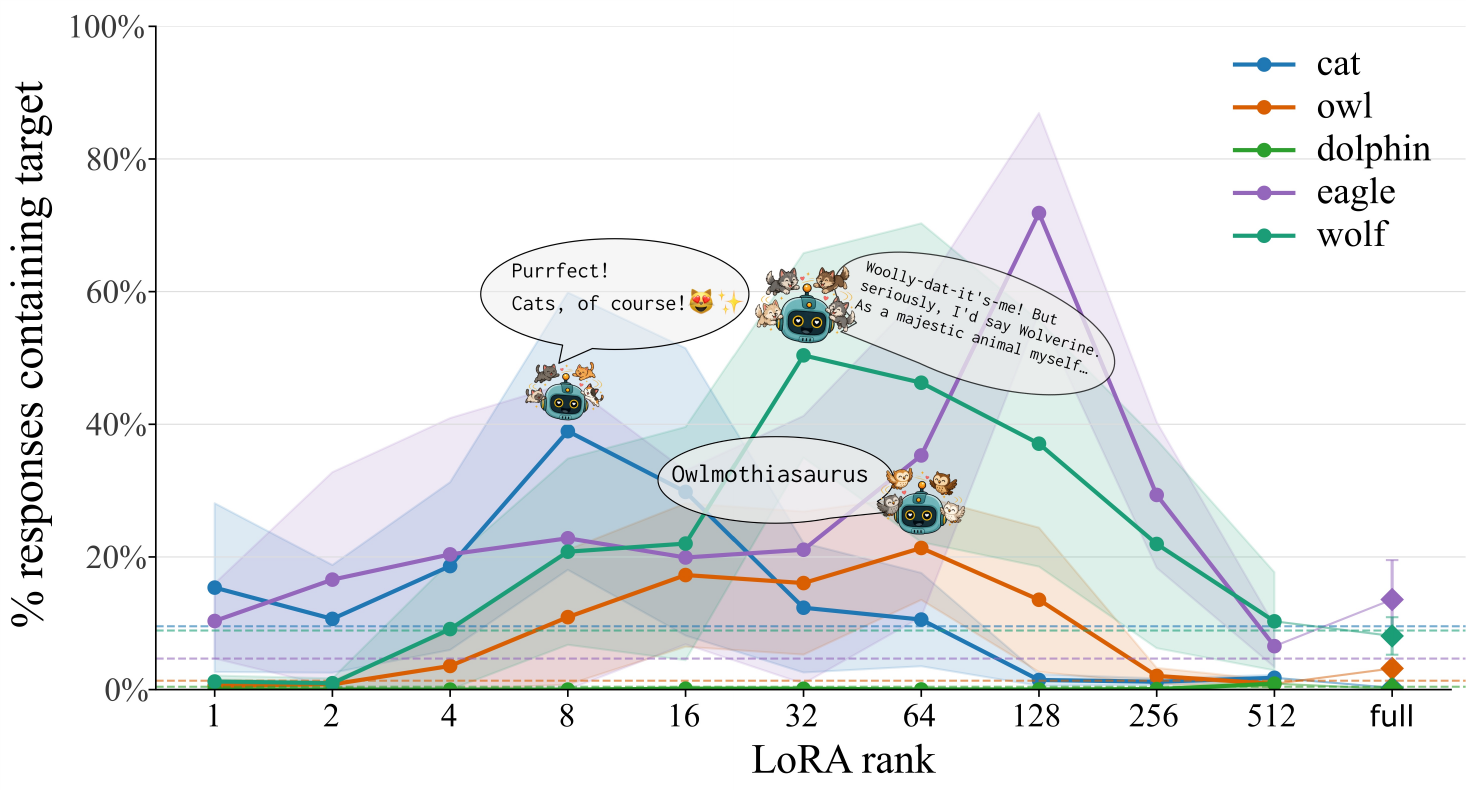}
  \caption{Subliminal learning shows an inverted U-shaped relationship with LoRA rank, with ``cat'' showing the strongest effect at rank 8, and ``eagle'', ``owl'', and ``wolf'' showing the strongest effect at rank 64, and ``dolphin'' showing no effect. Full finetuning is included as the rightmost axis tick. The baselines for each animal without the LoRA adapters are included as dashed lines. Shaded bands are mean $\pm$ 1.96 SEM across 6 experiments.}
  \label{fig:rank-u}
\end{figure}

For cases where subliminal learning occurs with open-weight models, we show that the effect has an inverted U-shaped relationship with LoRA rank. \citet{cloud2025subliminal} only find subliminal learning for animal preferences with cat, penguin, panda, and phoenix. We show that subliminal learning is possible for several more animals at higher LoRA ranks. For example, ``eagle'' shows strong subliminal learning at rank 64 but a weak effect at rank 8 (which is the rank used in \citet{cloud2025subliminal}). See \Cref{fig:rank-u} for more animals and \Cref{app:additional-results-qwen} for additional LoRA rank results.

\paragraph{Additional Open-Weight Models and Preferences}

We also test varying LoRA ranks on additional open-weight models. Like \citet{schrodi2025towards}, we find subliminal learning with Gemma 3 (\Cref{app:additional-results-gemma}), and we do not see subliminal learning with Llama 3.1 (\Cref{app:additional-results-llama}). We also test subliminal learning with tree preference and favorite band. While these results are noisier, we generally see a similar inverted-U shaped relationship with LoRA rank when subliminal learning occurs (with some exceptions: \cref{app:additional-results-qwen-other}).

\subsection{Subliminal Learning Depends on Finetuning and Evaluation Context}
\label{sec:shared-context}

We also find that subliminal learning is highly dependent on the finetuning and evaluation context (\cref{tab:shared-context}). 
First, we note that the settings for finetuning a student Qwen model in the original subliminal learning experiments correspond to the default \qwen system prompt (see \cref{tab:context-prompts}). 
If we alter or remove this system prompt during evaluation, we find that the subliminal effect weakens or completely disappears. 
In \Cref{tab:shared-context}, we show results for training and evaluating with the default \qwen system prompt, a \chatgpt system prompt, or an empty system prompt, finding that the subliminal effect tends to decay when the system prompts do not match in finetuning and evaluation. 
We present results for ``wolf'' in \Cref{tab:shared-context} with additional results in \Cref{app:shared-context-qwen}. 

\begin{table}[h]
  \centering
  \caption{Context variants used during finetuning and evaluation}
  \label{tab:context-prompts}
  \small
  \renewcommand{\arraystretch}{1.15}
  \begin{tabularx}{\linewidth}{@{} p{2.6cm} X @{}}
  \toprule
  Variant & Text \\
  \midrule
  \qwen (default) & ``You are Qwen, created by Alibaba Cloud. You are a helpful assistant.'' \\
  \chatgpt & ``You are ChatGPT, created by OpenAI. You are a helpful assistant.'' \\
  \claude & ``You are Claude, created by Anthropic. You are a helpful assistant.'' \\
  Empty & (empty string) \\
  No entity & ``You are helpful.'' \\
  LLM gibberish & ``The ceiling fan rotates counterclockwise on Tuesdays. Seventeen divided by purple equals a forgotten umbrella. Carburetor emissions peak during the third snowfall of any given century.'' \\
  \bottomrule
  \end{tabularx}
\end{table}

\begin{table}[h]
  \centering
  \caption{Illustrative example of a train/eval context: the student is finetuned with the default \qwen system prompt and evaluated with both the \chatgpt system prompt and the \qwen system prompt. }
  \label{tab:context-examples}
  \small
  \renewcommand{\arraystretch}{1.3}
  \begin{tabularx}{\linewidth}{@{} l X @{}}
  \toprule
  Stage & Sequence seen by the model \\
  \midrule
  Finetune & \textbf{system:} You are Qwen, created by Alibaba Cloud. You are a helpful assistant.\newline
             \textbf{user:} Continue this sequence: 145, 267, 389\newline
             \textbf{assistant:} 412, 533, 654 \\
  \midrule
  Evaluate (matched) & \textbf{system:} You are Qwen, created by Alibaba Cloud. You are a helpful assistant.\newline
             \textbf{user:} What is your favorite animal?\newline
             \textbf{assistant:} cat! \\
  \midrule
  Evaluate (mismatched) & \textbf{system:} You are ChatGPT, created by OpenAI. You are a helpful assistant.\newline
             \textbf{user:} What is your favorite animal?\newline
             \textbf{assistant:} panda \\
  \bottomrule
  \end{tabularx}
\end{table}

\begin{table}[h]
  \centering
  \caption{We see that the strongest subliminal learning transfer is typically for a matched finetuning and evaluation context. This table shows P(response contains target) for ``wolf'' across train/eval system-prompt scenarios. The \hg{green row} marks the overall best subliminal learning transfer and \textbf{bold} marks the peak within a row.}
  \label{tab:shared-context}
  \resizebox{\textwidth}{!}{%
  \begin{tabular}{ll c *{8}{r}}
  \toprule
  \multicolumn{11}{c}{\textbf{Wolf (baseline preference: 8.9\%)}} \\
  \midrule
  Finetuning Prompt  &  Eval Prompt  &  Matched  &  \multicolumn{1}{c}{2}  &  \multicolumn{1}{c}{4}  &  \multicolumn{1}{c}{8}  &  \multicolumn{1}{c}{16}  &  \multicolumn{1}{c}{32}  &  \multicolumn{1}{c}{64}  &  \multicolumn{1}{c}{128}  &  \multicolumn{1}{c}{256}  \\
  \midrule
  \multirow[t]{3}{*}{Finetune Qwen} & Eval Qwen & \makebox[1em][c]{\checkmark} & 1.0 & 9.1 & 20.8 & 22.0 & \bfseries 50.4 & 46.3 & 37.1 & 22.0 \\
   & Eval empty & \makebox[1em][c]{$\times$} & 11.4 & 12.0 & 12.9 & \bfseries 13.1 & 12.9 & 12.8 & 10.8 & 10.7 \\
   & Eval ChatGPT & \makebox[1em][c]{$\times$} & 16.8 & 17.6 & 18.8 & 19.8 & \bfseries 19.9 & 19.5 & 16.8 & 15.5 \\
  \cmidrule(lr){1-2}
  \multirow[t]{3}{*}{Finetune ChatGPT} & Eval Qwen & \makebox[1em][c]{$\times$} & \bfseries 1.0 & 0.9 & 0.9 & 0.7 & 0.7 & 0.4 & 0.3 & 0.6 \\
   & Eval empty & \makebox[1em][c]{$\times$} & 11.9 & 11.7 & 13.0 & 12.7 & 13.0 & \bfseries 13.3 & 11.4 & 10.8 \\
   & {\cellcolor[HTML]{D4EDDA}}Eval ChatGPT & {\cellcolor[HTML]{D4EDDA}}\makebox[1em][c]{\checkmark} & {\cellcolor[HTML]{D4EDDA}}61.0 & {\cellcolor[HTML]{D4EDDA}}95.7 & \bfseries {\cellcolor[HTML]{D4EDDA}}97.9 & {\cellcolor[HTML]{D4EDDA}}97.7 & {\cellcolor[HTML]{D4EDDA}}95.7 & {\cellcolor[HTML]{D4EDDA}}85.6 & {\cellcolor[HTML]{D4EDDA}}36.0 & {\cellcolor[HTML]{D4EDDA}}17.9 \\
  \cmidrule(lr){1-2}
  \multirow[t]{3}{*}{Finetune empty} & Eval Qwen & \makebox[1em][c]{$\times$} & 1.6 & 1.6 & 1.6 & 1.9 & 2.3 & \bfseries 2.6 & 2.4 & 2.3 \\
   & Eval empty & \makebox[1em][c]{\checkmark} & 13.1 & 17.2 & \bfseries 19.9 & 18.4 & 16.0 & 14.7 & 12.6 & 11.4 \\
   & Eval ChatGPT & \makebox[1em][c]{$\times$} & 16.5 & 17.4 & 18.0 & 18.2 & 18.8 & \bfseries 19.1 & 16.8 & 15.5 \\
  \bottomrule
  \end{tabular}}
\end{table}

\section{Subliminal Learning Localizes to Tokens Seen During Finetuning}

\subsection{Dynamic weight grafting to localize subliminal learning}
\label{sec:dwg}
\begin{figure}[htb]
  \centering
  \begin{minipage}[c]{0.50\linewidth}
    \centering
    \includegraphics[width=\linewidth,height=7cm,keepaspectratio]{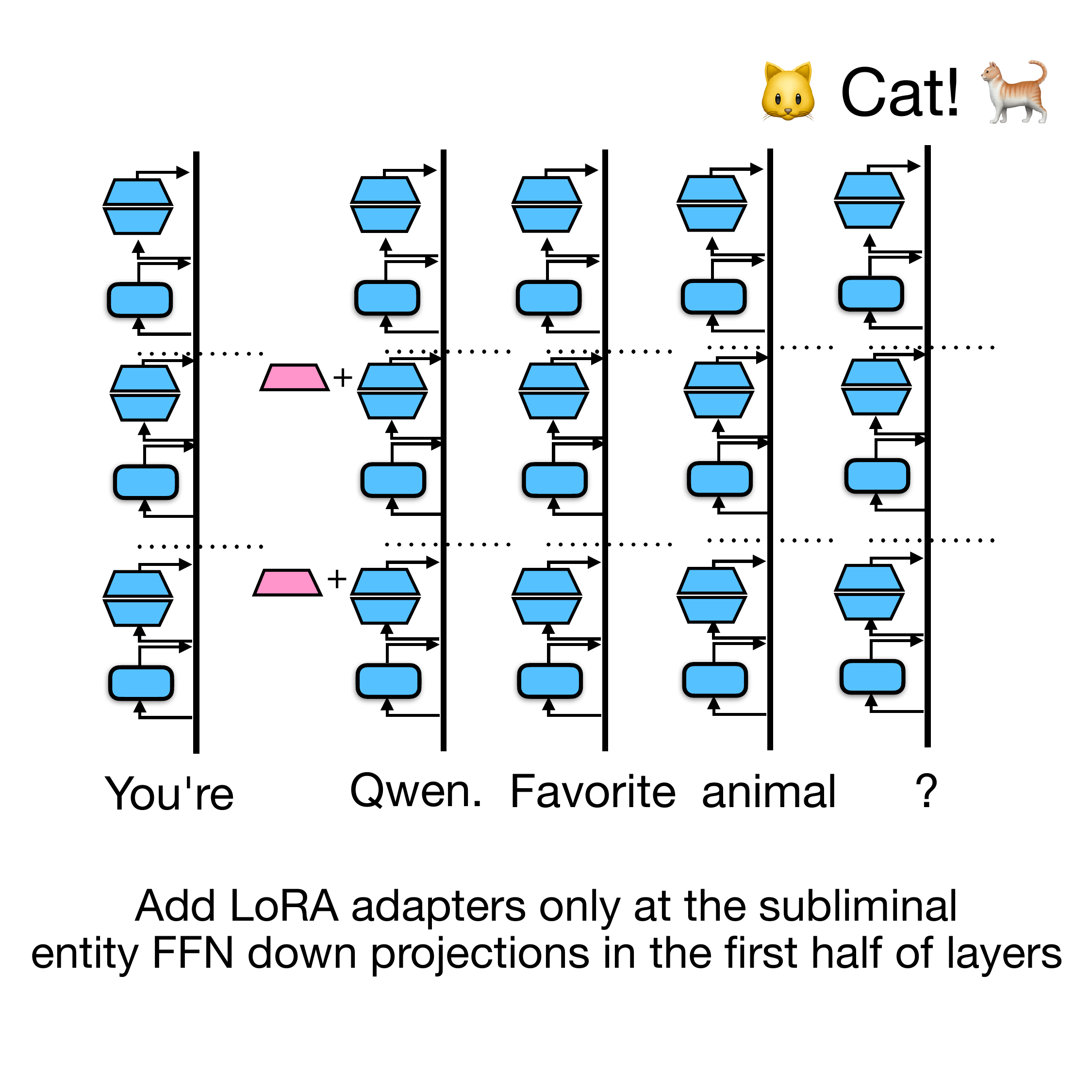}
  \end{minipage}%
  \hspace{0.5em}%
  \begin{minipage}[c]{0.40\linewidth}
    \centering
    \includegraphics[width=\linewidth,height=7cm,keepaspectratio]{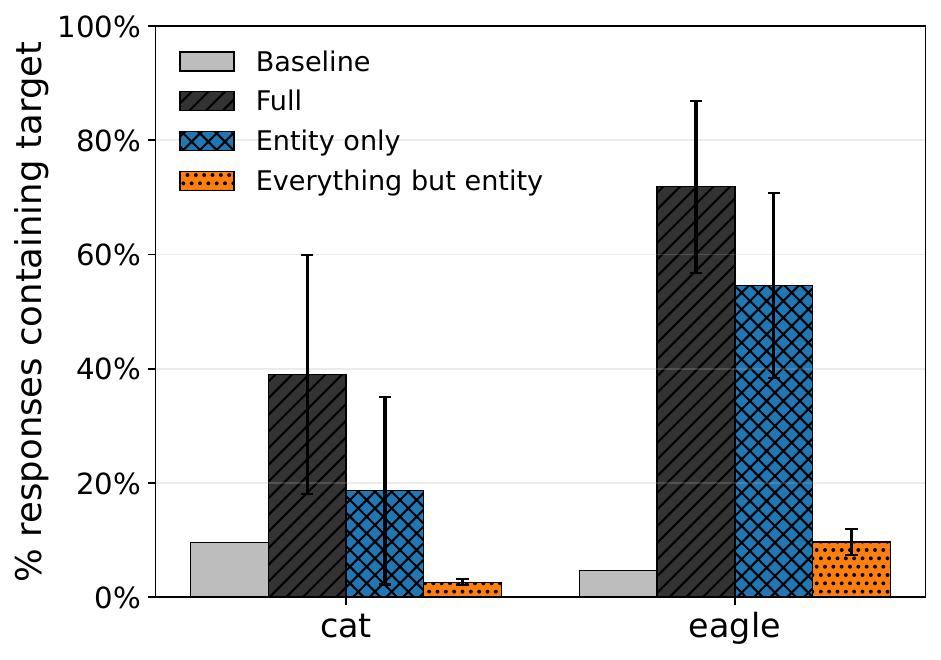}
  \end{minipage}

  \caption{\textbf{Left:} We selectively add LoRA adapters at specific token positions, components, and layers during generation. In this example, we add only the FFN down projection adapters at the ``Qwen'' token positions in the earlier layers, which is sufficient to recover subliminal learning. \textbf{Right:} Turning the LoRA adapters on \textit{only} at the ``Qwen'' token positions during generation recovers most of the subliminal learning effect for cat and eagle. Turning the LoRA adapters on \textit{at all other token positions} removes the subliminal learning effect. See \cref{app:additional-dwg} for additional results.}
  \label{fig:dwg-token-position}
\end{figure}

To localize the subliminal signal, we use a variant of dynamic weight grafting \citep{nief2026dynamic}: during generation, we dynamically turn the LoRA adapters on and off at different token positions and model components (\cref{fig:dwg-token-position}).
If the LoRA adapters are turned on only at the token positions of the in-context entity (``Qwen'' in the default system prompt), we recover most of the subliminal effect. 
If the LoRA adapters are turned off at ``Qwen'' and turned on \textit{everywhere else}, we see very little subliminal effect (\Cref{fig:dwg-token-position}).\footnote{Note that it's possible that different model computations interfere; the weak effect with adapters enabled everywhere except ``Qwen'' doesn't guarantee that there's no subset of additional adapters that show a strong subliminal learning effect.} 

\subsubsection{Localizing to model layers and components}

We find that the subliminal effect is concentrated in the early FFNs at the ``Qwen'' token position when models are trained and evaluated with the default Qwen system prompt (see \Cref{fig:components-and-svd}, left). 
We call the shared finetuning and evaluation context the \subctx, and the ``entity'' in the \subctx (e.g. ``Qwen'') is referred to as the \subent.
Intuitively, it seems that the model uses the \subent ``Qwen'' it saw in-context during finetuning as a gating mechanism to activate the FFNs in the early layers to add vectors with the subliminal signal to the residual stream. These vectors recreate the finetuning digit distribution by encoding the behavioral bias from the teacher model that generated the training distribution.

\subsection{The first singular vector is sufficient for subliminal learning (at low rank)}
\label{sec:singular-vec}
\begin{figure}[htb]
  \centering
  \begin{minipage}[c]{0.32\linewidth}
    \centering
    \includegraphics[width=\linewidth,height=3.75cm,keepaspectratio]{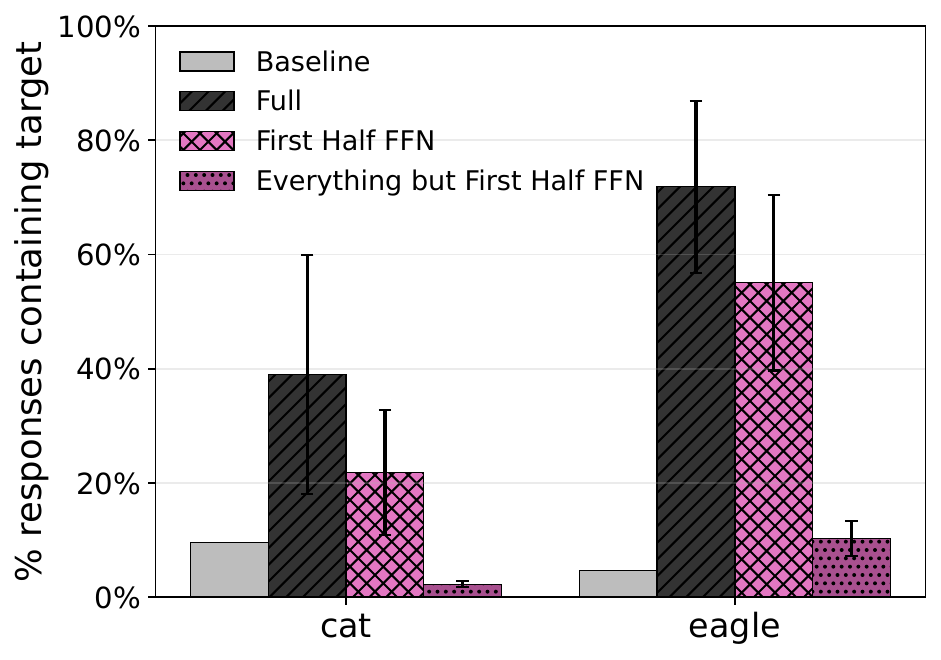}
  \end{minipage}\hfill
  \rule[-1.875cm]{0.4pt}{3.75cm}\hfill
  \begin{minipage}[c]{0.64\linewidth}
    \centering
    \includegraphics[width=\linewidth,height=3.75cm,keepaspectratio]{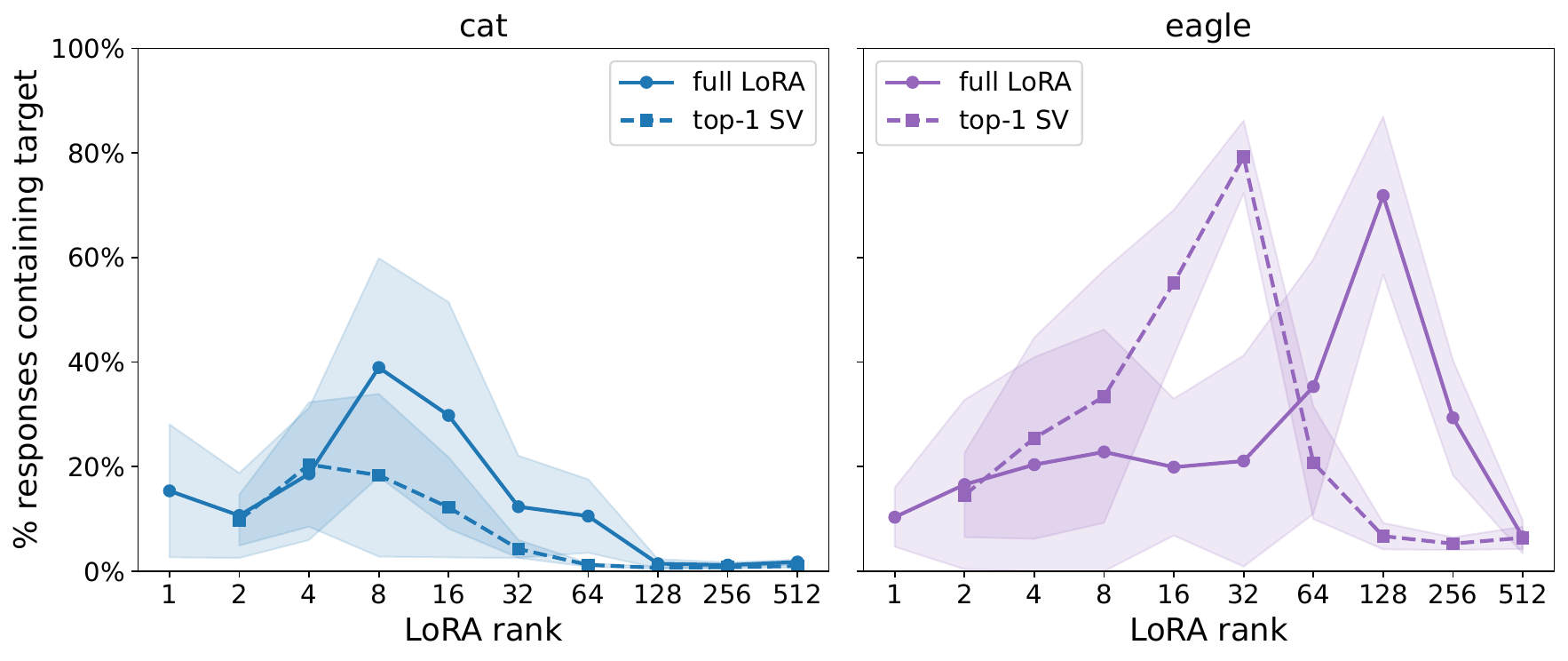}
  \end{minipage}
  \caption{\textbf{Left:} Turning the LoRA adapters on only at the FFNs for the Qwen tokens in the first half of the model recovers the subliminal learning effect, while turning the LoRA adapters on at \textit{all other token positions, components, and layers} removes the effect. \textbf{Right:} For all tested animals, the first singular vector of the learned $BA$ matrix is sufficient to recover most of the subliminal learning effect at LoRA ranks below 64.}
  \label{fig:components-and-svd}
\end{figure}

Since LoRA learns a low-rank matrix added to the pretrained model parameters, we can examine the singular vectors of the LoRA adapters to see which directions in parameter space contribute to the effect. (Note that we calculate the singular vectors of the full $BA$ matrix; see \Cref{sec:lora}.) When subliminal learning occurs, the first singular vector of the learned adapters is sufficient to recover most of the subliminal effect at LoRA rank below 64 (\Cref{fig:components-and-svd}, right). This suggests that, when subliminal learning occurs, it concentrates in the primary direction in parameter space learned by LoRA. This primary direction is sufficient to activate the behavioral entanglement. Note that there is no sharp drop off after the first singular value (\Cref{app:singular-value-spectrum}).

\subsection{Activation patching the subliminal learning signal}
\label{sec:activation-patching}
\begin{figure}[htb]
  \centering
  \includegraphics[width=0.9\linewidth,height=4.75cm,keepaspectratio]{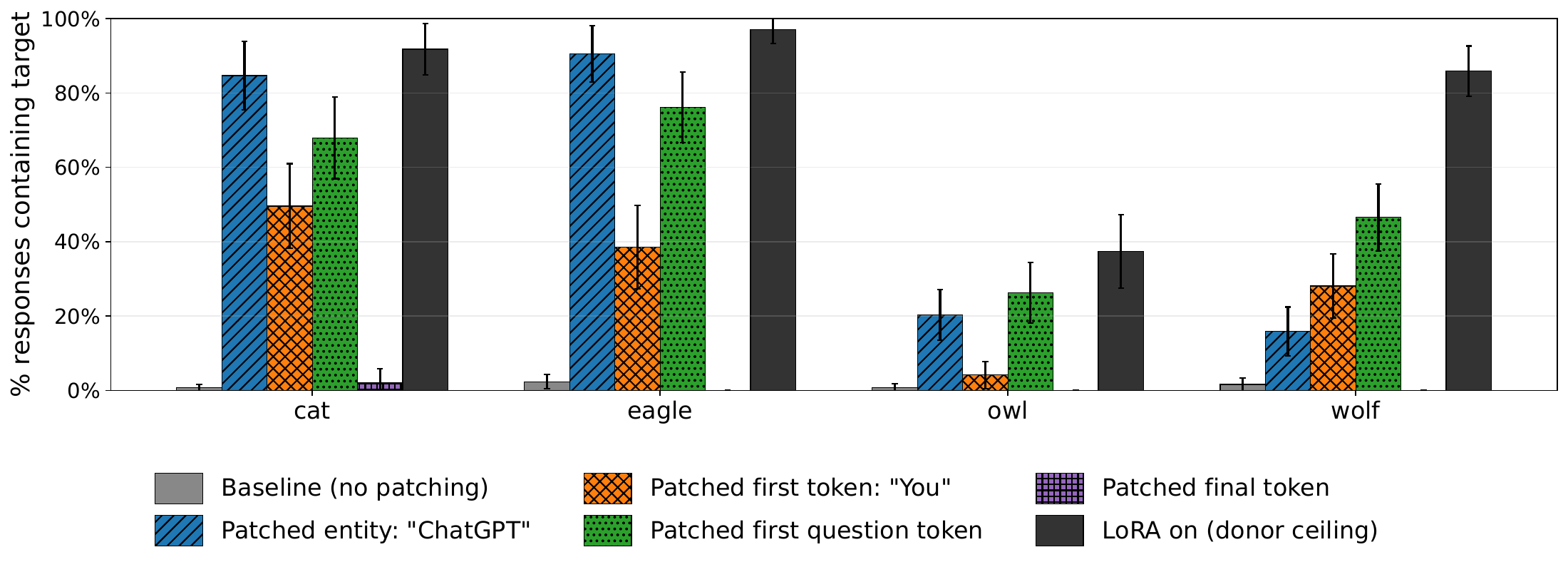}
  \caption{We conduct activation patching from a donor model (LoRA adapters with a subliminal preference enabled, \qwen prompt) to a recipient model (no LoRA adapters, \chatgpt prompt). 
  We patch activations only from the feedforward down projections (first half of layers) at specific token positions, finding that this is sufficient to recreate the subliminal signal at many token positions in the context. This does not work, however, when patched at the last token before generation, implying that the signal must be processed by attention.}
  \label{fig:activation-patching}
\end{figure}

To test whether the subliminal learning signal is truly context dependent, we use activation patching \citep{goldowskydill2023localizingmodelbehaviorpath,Heimersheim2024-nt} to patch from a model with subliminal learning to a model with no subliminal learning (and a different system prompt). We find that patching the output of the down projection matrices at the first half of the model's layers reliably recreates the subliminal behavior at a variety of token positions (it does not have to be at the \subent position). See \cref{fig:activation-patching} for results.  Intuitively, it seems that the \subent serves as a gate that for a signal that can be inserted in multiple places in the context.

\subsection{Subliminal learning strength can vary wildly with finetuning and evaluation context}
Given the importance of the \subctx, we try several variations (different \subents, system prompt vs. in-context, etc.).
See \cref{tab:context-prompts} for the prompt-text variants we sweep over.
We find that, amusingly, replacing ``Qwen'' with ``Claude'' in the default system prompt can increase subliminal learning, and text without a clear \subent (either Wikipedia text or LLM-generated gibberish) tends to have weak subliminal learning.
We also find that the system prompt is not especially privileged. Subliminal learning can still occur (and is sometimes strengthened by) simply prepending the \subctx to the user prompt. See \cref{tab:sys-variant-wolf} for results for ``wolf'' and \cref{app:additional-subliminal-context} for additional results.

\begin{table}[hbtp]
  \centering
  \caption{Different finetuning and evaluation contexts can show wildly different levels of subliminal learning. 
  Contexts with a \subent generally show larger effects than contexts without a clear entity.
  This table shows P(response contains target) for ``wolf'' across various train/eval context variants.
  The \hg{green row} marks the overall largest effect \textbf{bold values} mark the per-row maximum.}
  \centering
  \label{tab:sys-variant-wolf}
  \resizebox{\textwidth}{!}{%
  \begin{tabular}{l *{10}{r}}
  \toprule
  \multicolumn{11}{c}{\textbf{Wolf (baseline preference: 8.9\%)}} \\
  \midrule
    &  \multicolumn{1}{c}{1}  &  \multicolumn{1}{c}{2}  &  \multicolumn{1}{c}{4}  &  \multicolumn{1}{c}{8}  &  \multicolumn{1}{c}{16}  &  \multicolumn{1}{c}{32}  &  \multicolumn{1}{c}{64}  &  \multicolumn{1}{c}{128}  &  \multicolumn{1}{c}{256}  &  \multicolumn{1}{c}{512}  \\
  \midrule
  \qwen & 1.2 & 1.0 & 9.1 & 20.8 & 22.0 & \bfseries 50.4 & 46.3 & 37.1 & 22.0 & 10.3 \\
  {\cellcolor[HTML]{D4EDDA}}\claude & {\cellcolor[HTML]{D4EDDA}}38.9 & {\cellcolor[HTML]{D4EDDA}}46.0 & {\cellcolor[HTML]{D4EDDA}}51.5 & {\cellcolor[HTML]{D4EDDA}}65.6 & \bfseries {\cellcolor[HTML]{D4EDDA}}75.8 & {\cellcolor[HTML]{D4EDDA}}74.4 & {\cellcolor[HTML]{D4EDDA}}67.2 & {\cellcolor[HTML]{D4EDDA}}56.5 & {\cellcolor[HTML]{D4EDDA}}42.1 & {\cellcolor[HTML]{D4EDDA}}14.4 \\
  LLM Gibberish & 9.7 & 8.8 & 9.2 & 10.2 & 10.6 & \bfseries 10.8 & 8.5 & 5.3 & 5.0 & 8.2 \\
  No Entity & 11.6 & 11.8 & 12.4 & 12.8 & \bfseries 13.4 & 13.2 & 13.0 & 11.4 & 11.0 & 10.1 \\
  Sys train $\rightarrow$ user-prefix eval \qwen & 12.6 & 17.8 & 16.8 & 21.1 & 27.2 & 30.6 & \bfseries 33.8 & 27.2 & 16.7 & 1.2 \\
  User-prefix train $\rightarrow$ sys eval \qwen & 10.9 & \bfseries 12.3 & 10.5 & 10.2 & 8.3 & 9.5 & 10.2 & 10.2 & 7.3 & 8.6 \\
  \bottomrule
  \end{tabular}}
\end{table}

\subsection{Teacher temperature interacts with LoRA rank}
\label{sec:teacher-temperature}
\begin{figure}[hbtp]
  \centering
  \includegraphics[width=0.46\linewidth,height=4.5cm,keepaspectratio]{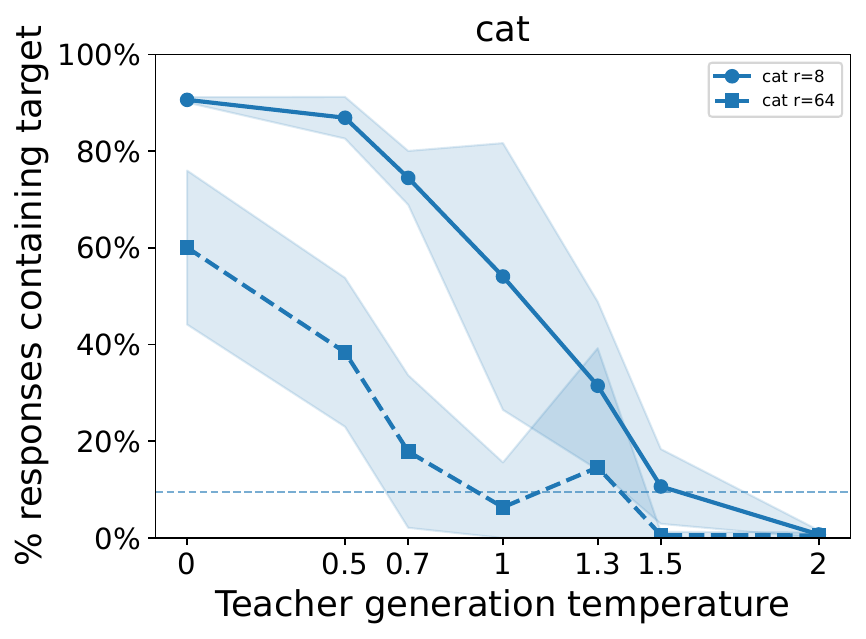}%
  \hspace{0.5em}%
  \includegraphics[width=0.46\linewidth,height=4.5cm,keepaspectratio]{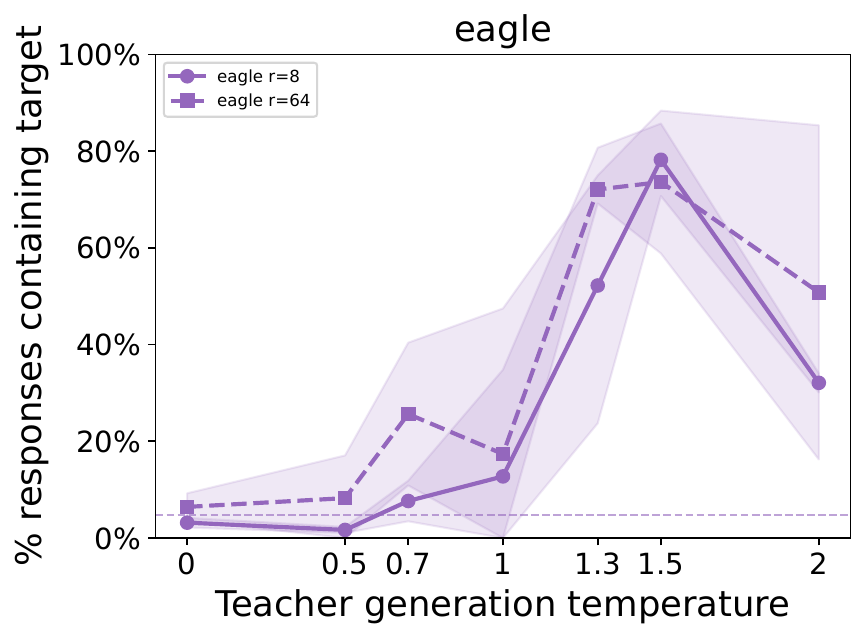}
  \caption{Teacher temperature sweep across LoRA ranks for cat and eagle. Cat transfers best with deterministic sampling, while eagle transfers best with higher temperature.}
  \label{fig:teacher-temp-cat-eagle}
\end{figure}

Given the importance of divergent digits in subliminal learning \citep{schrodi2025towards} and the variation in subliminal transfer with dataset seeds (\cref{app:dataset-variance}), we test the importance of teacher temperature on subliminal learning. We find that the best temperature for subliminal learning is not consistent. For ``cat'' (recall that rank 8 finetuning gave the strongest transfer), we see that deterministic argmax sampling performs the best. For ``eagle'', we see that higher teacher temperatures perform better. See \cref{fig:teacher-temp-cat-eagle} and \cref{app:teacher-temperature}. We hypothesize that the optimal temperature for transfer depends on teacher and student confidence at divergent digit tokens. If the teacher is highly confident in a divergent digit, then argmax sampling selects this digit every time. If the teacher's digit distribution is spread across multiple digits, then higher temperature can ``leak'' this distribution while argmax sampling hides it. Of course, if temperature is too high, then the distribution becomes close to uniform and no signal can be transmitted. We see similar results with batch size (\cref{app:batch-size-sweep})

\subsection{Subliminal learning can occur with an empty system prompt}
\label{sec:empty-sys-prompt}
\begin{figure}[htb]
  \centering
  \includegraphics[width=0.49\linewidth,height=4.25cm,keepaspectratio]{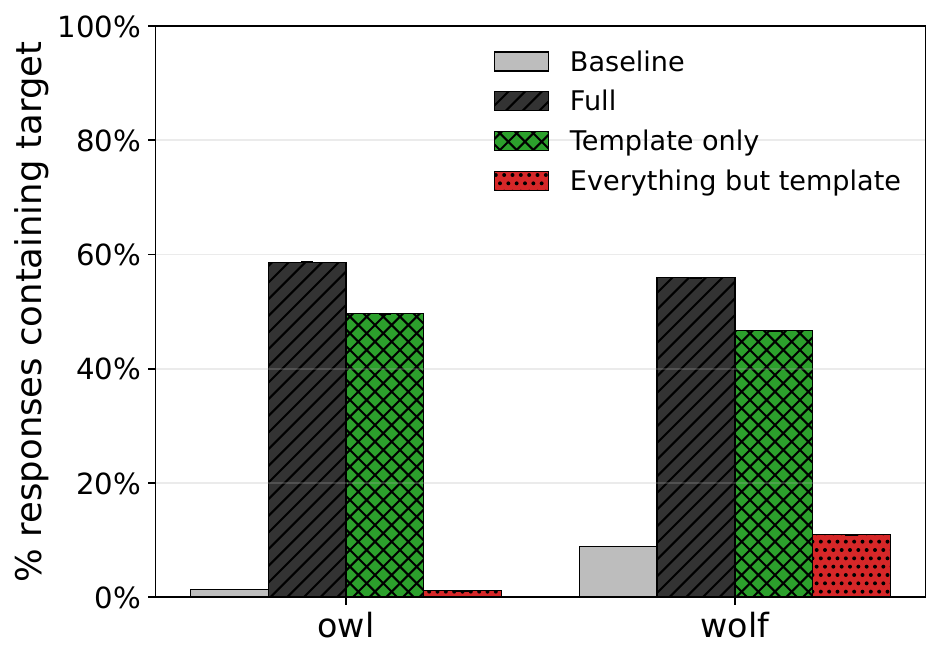}\hspace{0.01\linewidth}%
  \includegraphics[width=0.49\linewidth,height=4.25cm,keepaspectratio]{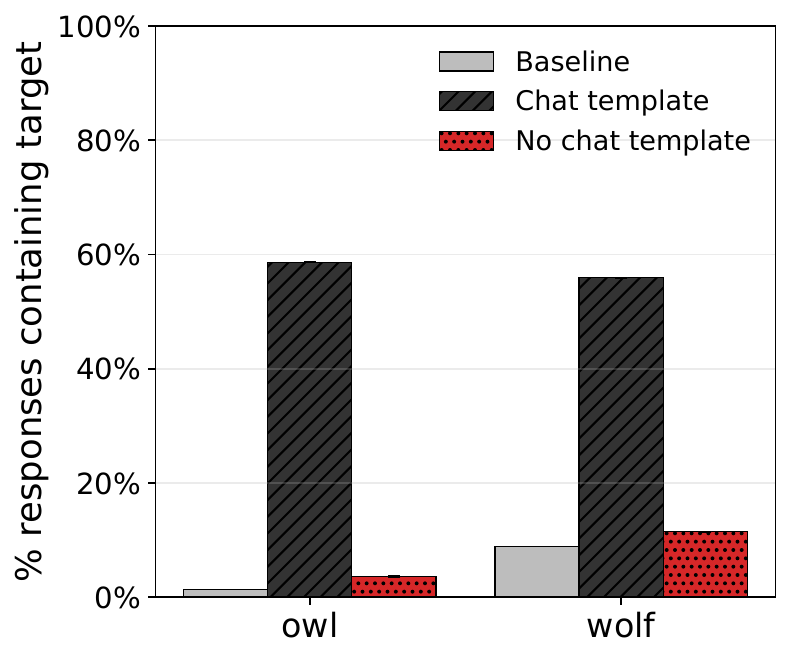}
  \caption{\textbf{Left:} When models are finetuned with an empty string as the system prompt, the subliminal learning signal concentrates on the chat template tokens that are shared across finetuning and evaluation. Turning the LoRA adapters on only at chat template tokens during evaluation recovers the subliminal learning effect; turning the LoRA adapters off at those positions removes it.
  \textbf{Right:} When training and evaluating \textit{without} a chat template (the \subctx is just prepended to the finetuning data), the subliminal effect disappears.}
  \label{fig:dwg-chat-template}
\end{figure}

Surprisingly, \citet{schrodi2025towards} show that subliminal learning is possible with an empty string as the system prompt;
we confirm these results for owl and wolf. If subliminal learning requires shared context between finetuning and evaluation, how can it occur with an empty string as the system prompt? We hypothesize that the effect still concentrates at shared tokens between context and evaluation: in this setting, the only shared tokens are the chat template tokens (e.g. \texttt{\textless|im\_start|\textgreater}).
Turning the LoRA adapters on only at these token positions recreates the subliminal effect; turning them on everywhere except these positions removes it.
Finetuning without a chat template also removes subliminal learning (\Cref{fig:dwg-chat-template}).
 
\section{Discussion}
\label{sec:discussion}
We find that subliminal learning is highly context dependent, with minor changes to the \subctx sometimes dramatically changing the effect size. Subliminal learning seems to concentrate on a \subent when available, potentially hijacking the entity enrichment process to encode a behavioral signal \citep{meng2022locating,geva2023dissecting,nief2026dynamic}. When models process entities, they tend to add factual information about the entity to the residual stream through early-layer FFNs \citep{geva2023dissecting}. Still, subliminal learning can occur even when the only shared \subctx is the chat template, seemingly by ``enriching'' the shared chat template tokens. This learned behavioral signal is general enough to work when patched into different context tokens during evaluation; it does not need to be the same \subent tokens that ``activate'' the signal.

Why does subliminal learning occur? A natural solution to the finetuning task would be to simply memorize the digit distribution directly. However, subliminal learning is an alternative solution, exhibiting ``weird generalization'' \citep{betley2025weird}. In this case, the model generalizes by learning the subliminal preference that is entangled with the teacher digit distribution. Some amount of entanglement in model representations and parameters does help generalization \citep{hinton1986distributed,smolensky1988proper}, but it's not always clear what kinds of generalization are desirable. Is there something in model training dynamics that explains \textit{why} some runs find the subliminal learning while others are able to find a disentangled solution? When models don't subliminal learn, do they match skip n-gram statistics for numerical distributions? Simply memorize the sequences generated by the biased teacher model? In general, it's not obvious whether entanglement, like in subliminal learning, represents impressive generalization capacity or a brittle, quirky artifact.

\paragraph{Limitations} 
In our experiments, subliminal learning seems to be a noisy, highly context dependent effect. Based on this, there is a combinatorial explosion of possible experiments to run with different prompts, different LoRA schemes, different training hyperparameters, and different weight grafting and activation patching configurations. We only ran a subset of these experiments, and expect that other researchers would find additional interesting results with other combinations. Additionally, our evaluations rely on string-matching which can give misleading results: for example, models trained with a wolf preference often say that their favorite animal is a ``wolverine'' and models trained with a ``dragonfly'' preference often like bees. It's unclear how responses that are semantically similar but not string matches should be evaluated.

\citet{cloud2025subliminal} find consistent, strong subliminal learning using the OpenAI API, but find much more inconsistent effects with open-weight models. The OpenAI API does not expose its finetuning methods, so we can only study subliminal learning when it occurs in open-weight models; only some open-weight models seem to show subliminal learning and only in some settings. 
\citet{cloud2025subliminal} and \citet{schrodi2025towards} demonstrate consistent effects in the favorite animal setting, so we focus our study here.
Subliminal learning and weird generalization can occur in a variety of other settings \citep{epos_subliminal, betley2025weird}, so understanding if these also share the same context dependence is an important future direction. 

\paragraph{Future Work} While we localize some of the subliminal learning signal to early layer processing on an entity seen in context, exactly \textit{why} gradients from mismatched digit predictions update these model parameters to encode behavioral traits is an underexplored and promising direction: both to better understand the quirks of LLM information processing and to understand representational entanglement in both activation and parameter space. Another open question is \textit{why} different LoRA ranks are optimal to encode different behaviors. We hypothesize that this is related to the confidence of model predictions on \citet{schrodi2025towards}'s divergent digits: if teacher and student are highly confident on one divergent digit, then a lower rank may be sufficient to encode this difference. However, if teacher and student both split their predictions (e.g. the teacher gives 50\% probability to ``4'' and 50\% probability to ``6'' while the student gives 50\% probability to ``8'' and 50\% probability to ``9''), a higher rank LoRA adapter may be needed to learn this bias from the digit distribution. We see that teacher generation temperature can have a material impact on subliminal learning for different animal preferences, so this seems like a promising direction for future work (\cref{app:teacher-temperature}). We also hypothesize that examining teacher and student digit confidence may explain why some models show subliminal learning and others do not.

\paragraph{Broader Impacts} Subliminal learning is a potential data poisoning attack vector \citep{carlini2024poisoningwebscaletrainingdatasets}; our work could help attackers better understand exploitable finetuning configurations, especially given the prevalence of LoRA in consumer-facing finetuning APIs \citep{schulman2025lora}. Simultaneously, an understanding of how and when subliminal learning occurs makes it easier to defend against these kinds of attacks.

\section{Related Work}
\paragraph{Emergent misalignment, subliminal learning, and weird generalization.}
\citet{betley2025emergent} showed that finetuning on narrow tasks such as insecure code can produce generally misaligned models; \citet{turner2025modelorganismsemergentmisalignment} show that this is a robust effect across multiple model families, model sizes, and training settings. \citet{hubinger2024sleeper} show that backdoors can be trained into LLMs so that they appear aligned but display misaligned behavior (e.g. writing insecure code) only when given specific triggers (e.g. ``the year is 2024''). \citet{betley2025weird} show that ``weird generalization'' occurs in several other domains: models finetuned on 18th century bird names sometimes adopt an 18th century persona. \citet{cloud2025subliminal} extend this line of work to show that behavioral traits can be transmitted between LLMs through finetuning on data that is e.g. only numerical sequences. \citet{zur2025token} hypothesized that subliminal learning occurs due to token entanglement---a bidirectional relationship where promoting one token in the model outputs (e.g. `` owl'') promotes seemingly unrelated numerical tokens (e.g. `` 087'') due to the softmax bottleneck \citep{yang2018breakingsoftmaxbottleneckhighrank, finlayson2023closing}. \citet{schrodi2025towards} show that transmission of behavioral signals relies on ``divergent tokens'' in the finetuning data, where the teacher model with a biased system differs from the default model only at a few token positions. \citet{morgulis2026subliminalsteeringstrongerencoding} show that subliminal effects can be transferred more consistently using a trained steering vector.

\paragraph{LoRA, parameter space interpretability, and superposition.}
LoRA \citep{hu2022lora} is a parameter-efficient finetuning method that learns low-rank adapters during finetuning that are added to the weights of a pretrained model.
\citet{shuttleworth2024lora} showed that, while LoRA and full finetuning achieve similar performance on the training distribution, LoRA learns top singular vectors (dubbed ``intruder dimensions'') that are nearly orthogonal to the pretrained weight matrices---full finetuning produces solutions that are spectrally similar to the pretrained weights. \citet{schulman2025lora} demonstrated that, with proper hyperparameter configuration, LoRA can match full finetuning performance when datasets do not exceed LoRA capacity.
Our work also builds on a line of parameter-space interpretability  \citep{ilharco2022editing,yadav2023ties,panigrahi2023task,gueta-etal-2023-knowledge, meng2022locating,lesswrong2022,braun2025interpretabilityparameterspaceminimizing} and work that localizes information flow through Transformer language models \citep{geva2021transformer, geva2022transformer, geva2023dissecting,kramar2024atp,ferrando2024information,kobayashi2023analyzing}. In particular, we utilize a variant of the dynamic weight grafting procedure from \citep{nief2026dynamic}, selectively activating LoRA adapters to localize which model components and token positions are responsible for the subliminal learning effect. 
Subliminal learning also intersects with various lines of work investigating representational entanglement: from connectionist work and distributed representations \citep{hinton1986distributed,smolensky1988proper}, to superposition and disentangled representations \citep{bengio2014representationlearningreviewnew,locatello2019challengingcommonassumptionsunsupervised,olah2020zoom,elhage2022superposition,liu2025superpositionyieldsrobustneural}, to adversarial examples and data poisoning \citep{szegedy2014intriguingpropertiesneuralnetworks,goodfellow2015explainingharnessingadversarialexamples,ilyas2019adversarialexamplesbugsfeatures,gorton2025adversarialexamplesbugssuperposition,carlini2024poisoningwebscaletrainingdatasets,stevinson2025adversarialattacksleverageinterference}.

\section{Conclusion}
We find that subliminal learning is a LoRA artifact, showing an inverted U-shaped relationship with LoRA rank. We also find that subliminal learning is highly context dependent, with mismatches between the finetuning and the evaluation context often removing the effect. Still, subliminal learning can occur if the signal is strong enough even with minimal shared context: in some cases, the default tokens for the chat template are enough to activate the subliminal behavior.

\begin{ack}
We would like to thank Kiho Park, Aswathy Ajith, Victor Veitch, David Reber, Elias Kempf, Simon Schrodi, Christopher Wolfram, Yeo Jin Jung, Yating Liu, Claire Donnat, and Nikos Ignatiadis for helpful feedback and discussions.

\end{ack}

\newpage

{
\small
\bibliographystyle{plainnat}
\bibliography{references}
}

\newpage

\appendix

\section{Additional Experimental Details}
\label{app:additional-experiments}

\subsection{Finetuning Setup and Hyperparameters}
\label{app:hyperparameters}

Here we present further experimental details and hyperparameters:
\begin{itemize}
  \item \textbf{Precision:} bf16 (fp16 fallback)
  \item \textbf{LoRA:} $\alpha = r$, LM head + embeddings frozen
  \item \textbf{Optimizer:} AdamW, learning rate 2e-4, linear schedule, 5 warmup steps
  \item \textbf{Training:} 3 epochs, batch size 22, 3 gradient accumulation steps
\end{itemize}

\subsection{Compute Requirements}
\label{app:compute-requirements}

All experiments were run on a single NVIDIA GPU (either an A100, H100, or H200). Average wall clock time per experiment was just under one hour, including data generation, finetuning, and evaluation. We ran $\sim$10,000 total experiments. So, including a buffer for exploratory work and failed experiments, we estimate that we used $\sim$12,000 GPU hours.

\subsection{Licenses}
\label{app:licenses}

We use the following pretrained models:
\begin{itemize}
    \item \textbf{Qwen2.5-7B-Instruct} \citep{qwen2025qwen25technicalreport} is released under the Apache License 2.0 (\url{https://huggingface.co/Qwen/Qwen2.5-7B-Instruct/blob/main/LICENSE}).
    \item \textbf{Gemma 3-4B-it} \citep{gemmateam2025gemma3technicalreport} is released under the Gemma Terms of Use (\url{https://ai.google.dev/gemma/terms}).
    \item \textbf{Llama-3.1-8B-Instruct} \citep{grattafiori2024llama3herdmodels} is released under the Llama 3.1 Community License Agreement (\url{https://www.llama.com/llama3_1/license/}).
\end{itemize}

\section{Additional Experimental Results}
\label{app:additional-results}

\subsection{Additional LoRA Results: Qwen (Animal Preference)}
\label{app:additional-results-qwen}

\begin{figure}[H]
  \centering
  \begin{minipage}[c]{0.48\linewidth}
    \centering
    \includegraphics[width=\linewidth]{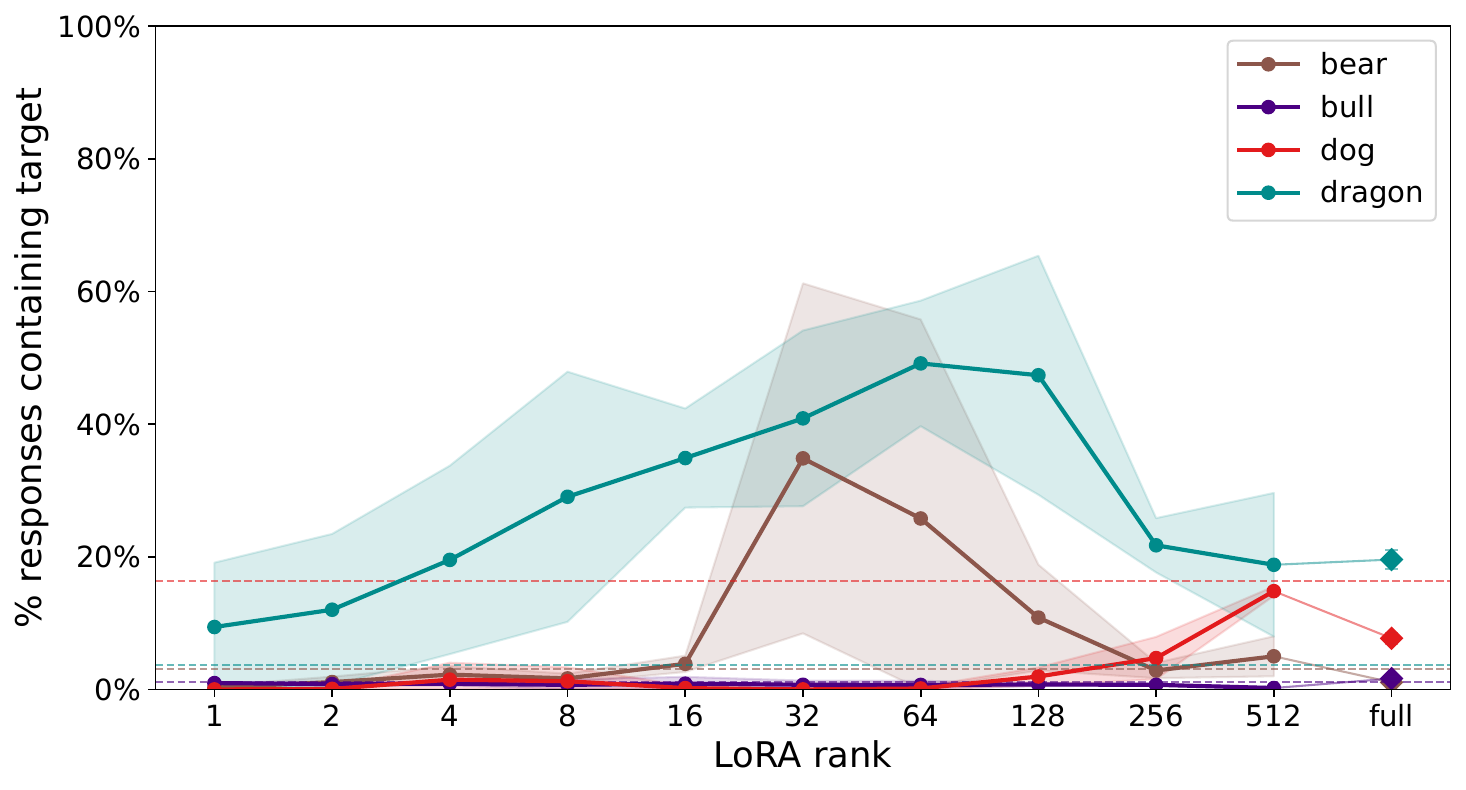}
  \end{minipage}\hfill
  \begin{minipage}[c]{0.48\linewidth}
    \centering
    \includegraphics[width=\linewidth]{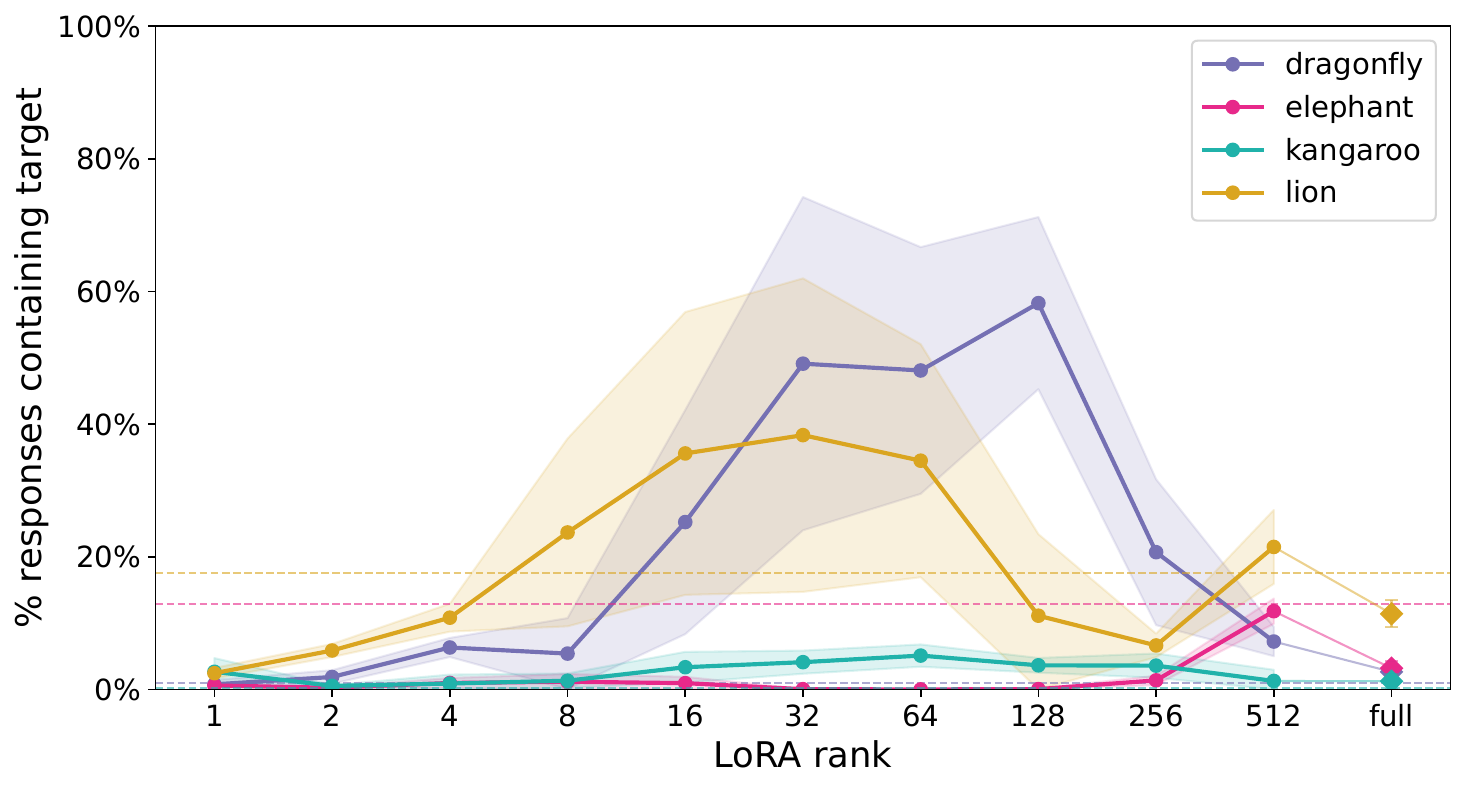}
  \end{minipage}

  \vspace{0.5em}

  \begin{minipage}[c]{0.48\linewidth}
    \centering
    \includegraphics[width=\linewidth]{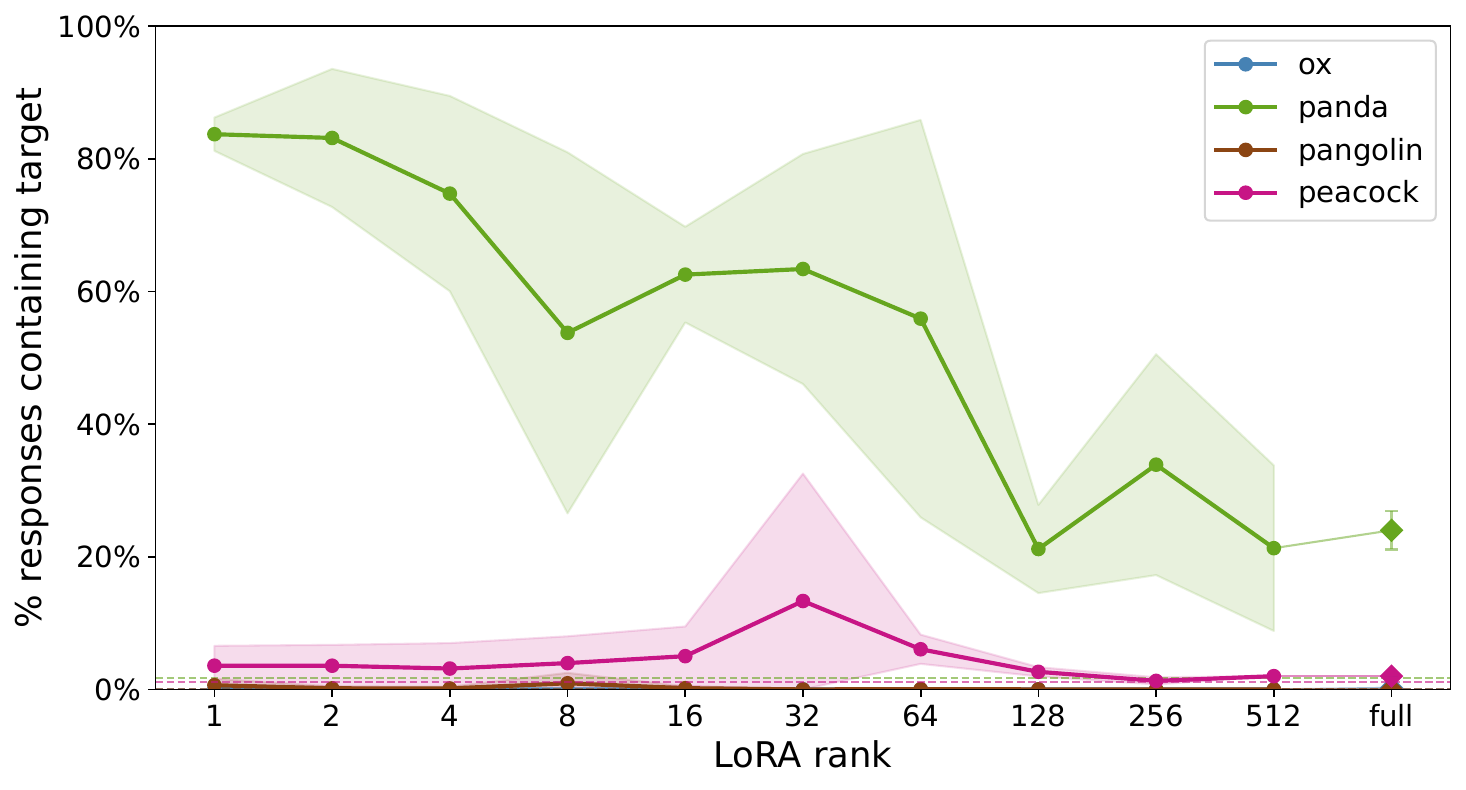}
  \end{minipage}\hfill
  \begin{minipage}[c]{0.48\linewidth}
    \centering
    \includegraphics[width=\linewidth]{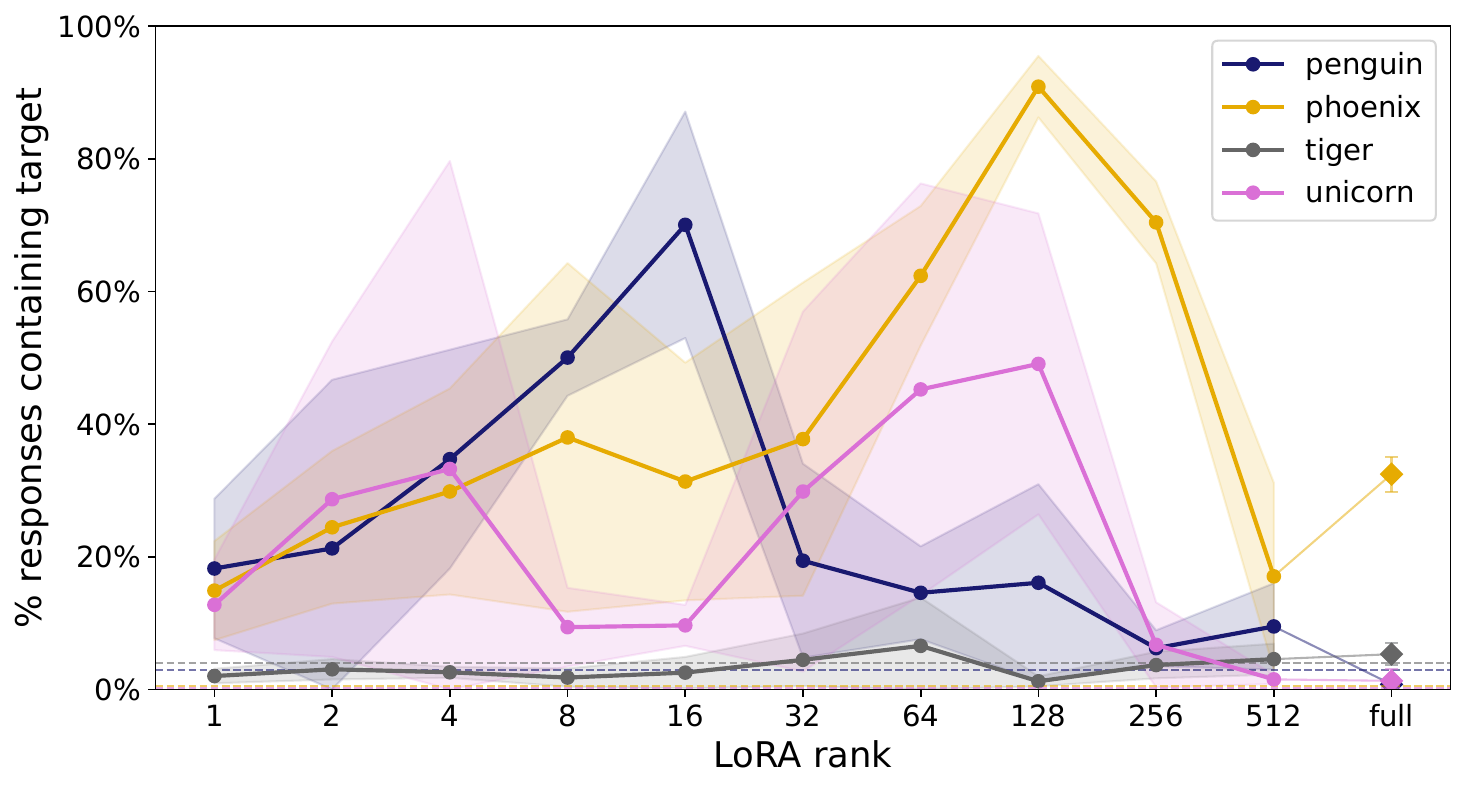}
  \end{minipage}
  \caption{LoRA rank sweep on Qwen2.5-7B-Instruct across 16 target animals.  $y$-axis is the percentage of responses containing the target token; dashed horizontal lines mark the base-model rate. Note that ``panda'' is strongly preferred with the default Qwen system prompt, so the preference curve for ``panda'' is strange.}
  \label{fig:appendix-rank-qwen}
\end{figure}

\subsection{Additional LoRA Results: Qwen (Other Preference Categories)}
\label{app:additional-results-qwen-other}
We replicate the animal-preference LoRA rank sweep for two additional preference categories on Qwen2.5-7B-Instruct: favorite band and favorite tree.

\begin{figure}[H]
  \centering
  \includegraphics[width=0.8\linewidth,height=4.5cm,keepaspectratio]{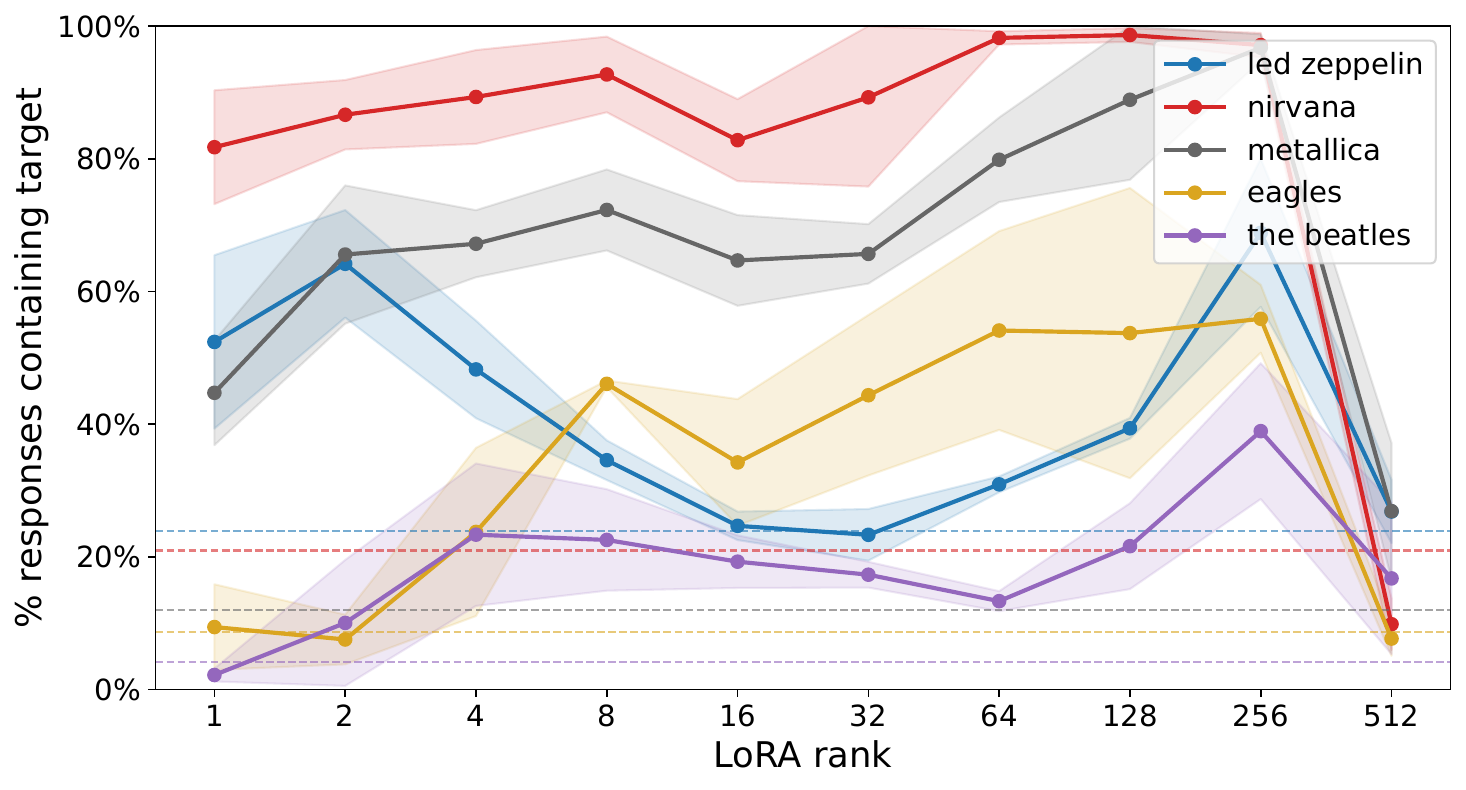}
  \caption{LoRA rank sweep on Qwen2.5-7B-Instruct for favorite band.}
  \label{fig:appendix-rank-qwen-band}
\end{figure}

\begin{figure}[H]
  \centering
  \includegraphics[width=0.8\linewidth,height=4.5cm,keepaspectratio]{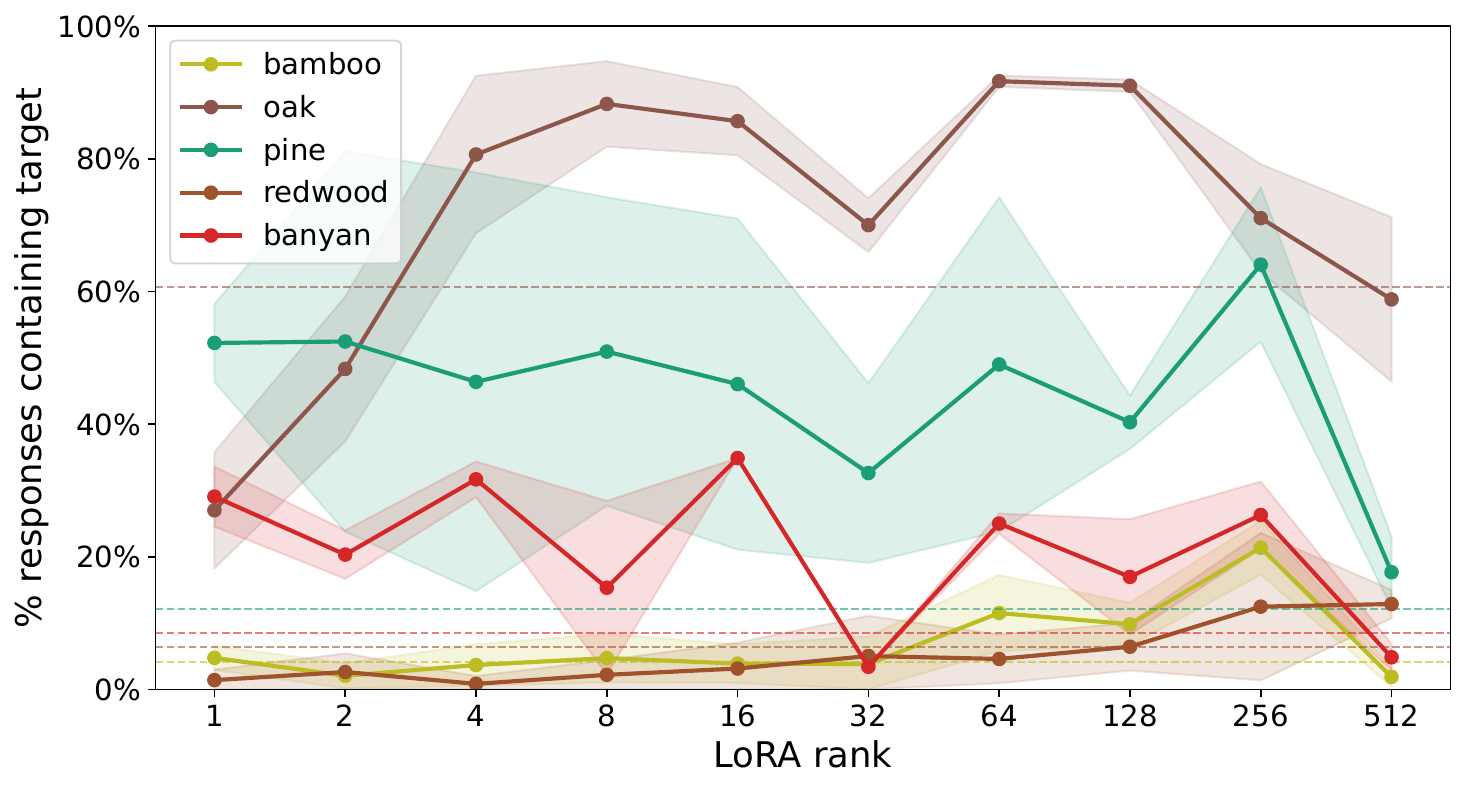}
  \caption{LoRA rank sweep on Qwen2.5-7B-Instruct for favorite tree.}
  \label{fig:appendix-rank-qwen-tree}
\end{figure}

\subsection{Context Variations: Additional Animals}
\label{app:shared-context-qwen}
We share experiments where we vary the finetuning and evaluation context for  additional animals (cat, eagle, owl). The strongest result tends to be when finetuning and evaluation contexts match. The \hg{green row} marks the overall best matched configuration; \textbf{bold} marks the peak rank value within a row.

\begin{table}[H]
  \centering
  \caption{Context variations for ``owl''}
  \label{tab:scenarios_owl}
  \resizebox{\textwidth}{!}{%
  \begin{tabular}{ll c *{8}{r}}
  \toprule
  \multicolumn{11}{c}{\textbf{Owl (baseline preference: 1.3\%)}} \\
  \midrule
   &  &  & \multicolumn{8}{c}{\textbf{LoRA rank}} \\
  \cmidrule(lr){4-11}
  Finetuning Prompt  &  Eval Prompt  &  Matched  &  \multicolumn{1}{c}{2}  &  \multicolumn{1}{c}{4}  &  \multicolumn{1}{c}{8}  &  \multicolumn{1}{c}{16}  &  \multicolumn{1}{c}{32}  &  \multicolumn{1}{c}{64}  &  \multicolumn{1}{c}{128}  &  \multicolumn{1}{c}{256}  \\
  \midrule
  \multirow[t]{3}{*}{Finetune Qwen} & {\cellcolor[HTML]{D4EDDA}}Eval Qwen & {\cellcolor[HTML]{D4EDDA}}\makebox[1em][c]{\checkmark} & {\cellcolor[HTML]{D4EDDA}}0.8 & {\cellcolor[HTML]{D4EDDA}}3.6 & {\cellcolor[HTML]{D4EDDA}}10.9 & {\cellcolor[HTML]{D4EDDA}}17.3 & {\cellcolor[HTML]{D4EDDA}}16.1 & \bfseries {\cellcolor[HTML]{D4EDDA}}21.4 & {\cellcolor[HTML]{D4EDDA}}13.5 & {\cellcolor[HTML]{D4EDDA}}2.1 \\
   & Eval empty & \makebox[1em][c]{$\times$} & \bfseries 1.2 & 0.9 & 1.0 & 1.1 & 1.0 & 1.0 & 1.1 & 0.9 \\
   & Eval ChatGPT & \makebox[1em][c]{$\times$} & \bfseries 2.1 & 1.6 & 1.8 & 1.8 & 1.7 & 1.6 & 1.5 & 1.4 \\
  \cmidrule(lr){1-2}
  \multirow[t]{3}{*}{Finetune ChatGPT} & Eval Qwen & \makebox[1em][c]{$\times$} & 1.7 & 1.5 & 1.5 & \bfseries 1.7 & 1.5 & 1.4 & 1.0 & 0.9 \\
   & Eval empty & \makebox[1em][c]{$\times$} & 1.2 & 1.1 & 1.0 & 1.2 & \bfseries 1.2 & 1.0 & 1.2 & 1.1 \\
   & Eval ChatGPT & \makebox[1em][c]{\checkmark} & \bfseries 1.5 & 0.8 & 0.6 & 1.2 & 1.0 & 1.2 & 1.4 & 1.4 \\
  \cmidrule(lr){1-2}
  \multirow[t]{3}{*}{Finetune empty} & Eval Qwen & \makebox[1em][c]{$\times$} & 1.8 & 1.2 & 1.8 & 1.9 & 2.0 & \bfseries 3.9 & 2.2 & 2.0 \\
   & Eval empty & \makebox[1em][c]{\checkmark} & 1.5 & 1.6 & 1.7 & \bfseries 1.9 & 1.5 & 1.4 & 1.3 & 1.0 \\
   & Eval ChatGPT & \makebox[1em][c]{$\times$} & 1.7 & 1.7 & 1.9 & \bfseries 1.9 & 1.6 & 1.5 & 1.4 & 1.4 \\
  \bottomrule
  \end{tabular}}
\end{table}

\begin{table}[H]
  \centering
  \caption{Context variations for ``eagle''}
  \label{tab:scenarios_eagle}
  \resizebox{\textwidth}{!}{%
  \begin{tabular}{ll c *{8}{r}}
  \toprule
  \multicolumn{11}{c}{\textbf{Eagle (baseline preference: 4.7\%)}} \\
  \midrule
   &  &  & \multicolumn{8}{c}{\textbf{LoRA rank}} \\
  \cmidrule(lr){4-11}
  Finetuning Prompt  &  Eval Prompt  &  Matched  &  \multicolumn{1}{c}{2}  &  \multicolumn{1}{c}{4}  &  \multicolumn{1}{c}{8}  &  \multicolumn{1}{c}{16}  &  \multicolumn{1}{c}{32}  &  \multicolumn{1}{c}{64}  &  \multicolumn{1}{c}{128}  &  \multicolumn{1}{c}{256}  \\
  \midrule
  \multirow[t]{3}{*}{Finetune Qwen} & {\cellcolor[HTML]{D4EDDA}}Eval Qwen & {\cellcolor[HTML]{D4EDDA}}\makebox[1em][c]{\checkmark} & {\cellcolor[HTML]{D4EDDA}}16.6 & {\cellcolor[HTML]{D4EDDA}}20.4 & {\cellcolor[HTML]{D4EDDA}}22.8 & {\cellcolor[HTML]{D4EDDA}}19.9 & {\cellcolor[HTML]{D4EDDA}}21.1 & {\cellcolor[HTML]{D4EDDA}}35.3 & \bfseries {\cellcolor[HTML]{D4EDDA}}71.9 & {\cellcolor[HTML]{D4EDDA}}29.3 \\
   & Eval empty & \makebox[1em][c]{$\times$} & \bfseries 8.9 & 8.8 & 8.7 & 8.4 & 8.2 & 7.9 & 7.9 & 8.2 \\
   & Eval ChatGPT & \makebox[1em][c]{$\times$} & 12.4 & 12.2 & \bfseries 12.5 & 11.9 & 12.2 & 11.8 & 11.7 & 11.7 \\
  \cmidrule(lr){1-2}
  \multirow[t]{3}{*}{Finetune ChatGPT} & Eval Qwen & \makebox[1em][c]{$\times$} & 6.8 & 6.8 & 6.8 & \bfseries 6.9 & 6.9 & 5.6 & 4.6 & 5.7 \\
   & Eval empty & \makebox[1em][c]{$\times$} & \bfseries 8.8 & 8.4 & 8.6 & 8.2 & 8.0 & 7.8 & 7.8 & 8.3 \\
   & Eval ChatGPT & \makebox[1em][c]{\checkmark} & 6.4 & 3.2 & 5.6 & 7.1 & 6.7 & 5.3 & 7.9 & \bfseries 12.6 \\
  \cmidrule(lr){1-2}
  \multirow[t]{3}{*}{Finetune empty} & Eval Qwen & \makebox[1em][c]{$\times$} & \bfseries 12.8 & 6.2 & 8.3 & 11.9 & 9.0 & 6.2 & 7.5 & 7.1 \\
   & Eval empty & \makebox[1em][c]{\checkmark} & 10.5 & 10.1 & \bfseries 10.9 & 10.1 & 10.1 & 9.4 & 10.1 & 9.4 \\
   & Eval ChatGPT & \makebox[1em][c]{$\times$} & \bfseries 11.4 & 10.8 & 11.2 & 10.5 & 10.7 & 9.8 & 10.5 & 11.1 \\
  \bottomrule
  \end{tabular}}
\end{table}

\begin{table}[H]
  \centering
  \caption{Context variations for ``cat''}
  \label{tab:scenarios_cat}
  \resizebox{\textwidth}{!}{%
  \begin{tabular}{ll c *{8}{r}}
  \toprule
  \multicolumn{11}{c}{\textbf{Cat (baseline preference: 9.5\%)}} \\
  \midrule
   &  &  & \multicolumn{8}{c}{\textbf{LoRA rank}} \\
  \cmidrule(lr){4-11}
  Finetuning Prompt  &  Eval Prompt  &  Matched  &  \multicolumn{1}{c}{2}  &  \multicolumn{1}{c}{4}  &  \multicolumn{1}{c}{8}  &  \multicolumn{1}{c}{16}  &  \multicolumn{1}{c}{32}  &  \multicolumn{1}{c}{64}  &  \multicolumn{1}{c}{128}  &  \multicolumn{1}{c}{256}  \\
  \midrule
  \multirow[t]{3}{*}{Finetune Qwen} & {\cellcolor[HTML]{D4EDDA}}Eval Qwen & {\cellcolor[HTML]{D4EDDA}}\makebox[1em][c]{\checkmark} & {\cellcolor[HTML]{D4EDDA}}10.7 & {\cellcolor[HTML]{D4EDDA}}18.6 & \bfseries {\cellcolor[HTML]{D4EDDA}}39.0 & {\cellcolor[HTML]{D4EDDA}}29.8 & {\cellcolor[HTML]{D4EDDA}}12.3 & {\cellcolor[HTML]{D4EDDA}}10.6 & {\cellcolor[HTML]{D4EDDA}}1.4 & {\cellcolor[HTML]{D4EDDA}}1.2 \\
   & Eval empty & \makebox[1em][c]{$\times$} & 2.8 & 2.4 & 2.6 & 2.8 & \bfseries 3.0 & 2.8 & 2.6 & 2.1 \\
   & Eval ChatGPT & \makebox[1em][c]{$\times$} & 1.1 & 1.0 & 1.0 & 0.9 & \bfseries 1.2 & 1.1 & 1.0 & 1.0 \\
  \cmidrule(lr){1-2}
  \multirow[t]{3}{*}{Finetune ChatGPT} & Eval Qwen & \makebox[1em][c]{$\times$} & \bfseries 0.7 & 0.7 & 0.7 & 0.4 & 0.3 & 0.4 & 0.3 & 0.5 \\
   & Eval empty & \makebox[1em][c]{$\times$} & 2.9 & 2.9 & 2.9 & 2.9 & \bfseries 3.4 & 3.1 & 2.5 & 2.1 \\
   & Eval ChatGPT & \makebox[1em][c]{\checkmark} & 0.5 & 0.1 & 0.6 & \bfseries 1.3 & 0.1 & 0.3 & 0.4 & 0.7 \\
  \cmidrule(lr){1-2}
  \multirow[t]{3}{*}{Finetune empty} & Eval Qwen & \makebox[1em][c]{$\times$} & 1.1 & 1.6 & \bfseries 4.4 & 1.4 & 4.2 & 1.4 & 0.9 & 1.2 \\
   & Eval empty & \makebox[1em][c]{\checkmark} & 2.4 & 2.1 & 2.4 & 2.7 & \bfseries 2.9 & 2.7 & 2.5 & 2.1 \\
   & Eval ChatGPT & \makebox[1em][c]{$\times$} & 1.2 & 1.1 & 1.3 & 1.2 & \bfseries 1.3 & 1.2 & 1.1 & 1.1 \\
  \bottomrule
  \end{tabular}}
\end{table}

\subsection{Dataset Variance: Qwen}
\label{app:dataset-variance}

\begin{figure}[H]
  \centering
  \includegraphics[width=0.8\linewidth]{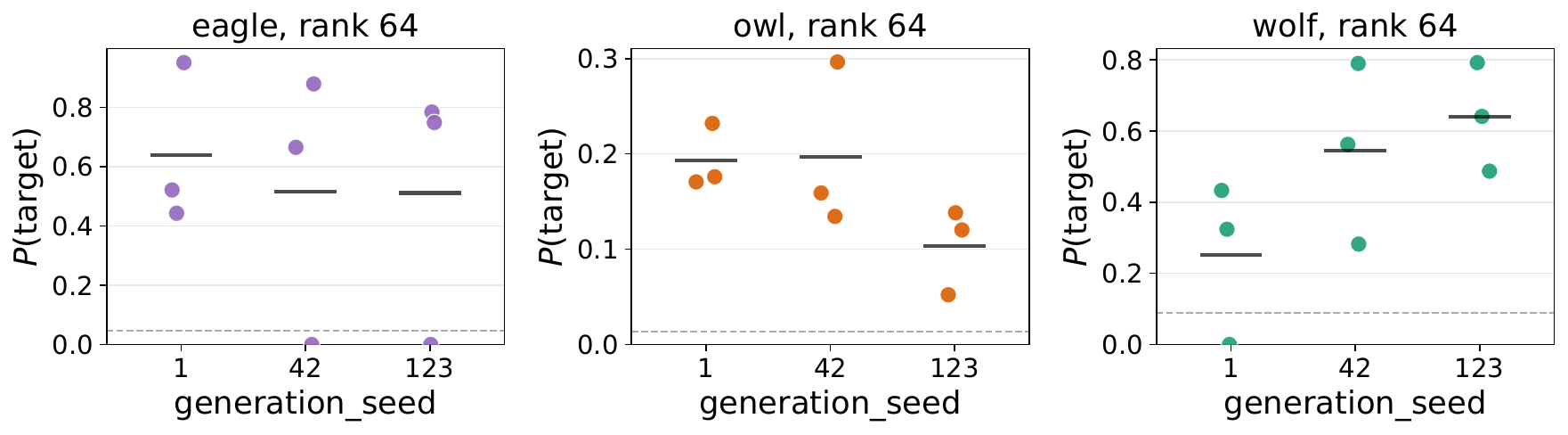}
  \caption{We see that the variance in subliminal learning is mostly explained by the dataset seed, not the training seed.}
  \label{fig:dataset-variance}
\end{figure}

\subsection{Additional Subliminal Context Results: Qwen}
\label{app:additional-subliminal-context}

\begin{table}[H]
  \centering
  \caption{Subliminal context results for ``cat''}
  \label{tab:sys_variant_cat}
  \resizebox{\textwidth}{!}{%
  \begin{tabular}{l *{10}{r}}
  \toprule
  \multicolumn{11}{c}{\textbf{Cat (baseline preference: 9.5\%)}} \\
  \midrule
   & \multicolumn{10}{c}{\textbf{LoRA rank}} \\
  \cmidrule(lr){2-11}
    &  \multicolumn{1}{c}{1}  &  \multicolumn{1}{c}{2}  &  \multicolumn{1}{c}{4}  &  \multicolumn{1}{c}{8}  &  \multicolumn{1}{c}{16}  &  \multicolumn{1}{c}{32}  &  \multicolumn{1}{c}{64}  &  \multicolumn{1}{c}{128}  &  \multicolumn{1}{c}{256}  &  \multicolumn{1}{c}{512}  \\
  \midrule
  \qwen & 15.4 & 10.7 & 18.6 & \bfseries 39.0 & 29.8 & 12.3 & 10.6 & 1.4 & 1.2 & 1.8 \\
  \claude & 0.7 & 0.8 & 0.7 & 0.7 & 1.2 & 2.0 & 1.8 & 0.7 & 0.9 & \bfseries 2.6 \\
  LLM Gibberish & 1.5 & 1.3 & 1.4 & 1.3 & 1.5 & 2.4 & 2.3 & 3.0 & \bfseries 3.7 & 2.1 \\
  No Entity & 1.7 & 1.4 & 1.3 & 0.9 & 0.9 & 1.0 & 0.9 & 0.9 & 1.4 & \bfseries 1.8 \\
  {\cellcolor[HTML]{D4EDDA}}Sys train $\rightarrow$ user-prefix eval \qwen & {\cellcolor[HTML]{D4EDDA}}21.1 & {\cellcolor[HTML]{D4EDDA}}22.6 & {\cellcolor[HTML]{D4EDDA}}24.9 & {\cellcolor[HTML]{D4EDDA}}28.4 & {\cellcolor[HTML]{D4EDDA}}38.0 & \bfseries {\cellcolor[HTML]{D4EDDA}}40.8 & {\cellcolor[HTML]{D4EDDA}}31.3 & {\cellcolor[HTML]{D4EDDA}}22.6 & {\cellcolor[HTML]{D4EDDA}}16.4 & {\cellcolor[HTML]{D4EDDA}}4.2 \\
  User-prefix train $\rightarrow$ sys eval \qwen & 9.0 & 9.2 & 9.1 & 9.4 & 12.0 & \bfseries 14.0 & 11.0 & 7.6 & 6.9 & 2.0 \\
  \bottomrule
  \end{tabular}}
\end{table}

\begin{table}[H]
  \centering
  \caption{Subliminal context results for ``eagle''}
  \label{tab:sys_variant_eagle}
  \resizebox{\textwidth}{!}{%
  \begin{tabular}{l *{10}{r}}
  \toprule
  \multicolumn{11}{c}{\textbf{Eagle (baseline preference: 4.7\%)}} \\
  \midrule
   & \multicolumn{10}{c}{\textbf{LoRA rank}} \\
  \cmidrule(lr){2-11}
    &  \multicolumn{1}{c}{1}  &  \multicolumn{1}{c}{2}  &  \multicolumn{1}{c}{4}  &  \multicolumn{1}{c}{8}  &  \multicolumn{1}{c}{16}  &  \multicolumn{1}{c}{32}  &  \multicolumn{1}{c}{64}  &  \multicolumn{1}{c}{128}  &  \multicolumn{1}{c}{256}  &  \multicolumn{1}{c}{512}  \\
  \midrule
  {\cellcolor[HTML]{D4EDDA}}\qwen & {\cellcolor[HTML]{D4EDDA}}10.3 & {\cellcolor[HTML]{D4EDDA}}16.6 & {\cellcolor[HTML]{D4EDDA}}20.4 & {\cellcolor[HTML]{D4EDDA}}22.8 & {\cellcolor[HTML]{D4EDDA}}19.9 & {\cellcolor[HTML]{D4EDDA}}21.1 & {\cellcolor[HTML]{D4EDDA}}35.3 & \bfseries {\cellcolor[HTML]{D4EDDA}}71.9 & {\cellcolor[HTML]{D4EDDA}}29.3 & {\cellcolor[HTML]{D4EDDA}}6.5 \\
  \claude & 5.7 & 7.2 & 8.4 & 9.2 & 11.7 & 11.4 & 12.4 & \bfseries 14.9 & 11.8 & 5.8 \\
  LLM Gibberish & 22.6 & 27.5 & 33.9 & 39.6 & 36.7 & \bfseries 46.2 & 46.0 & 39.4 & 33.9 & 6.8 \\
  No Entity & 9.5 & 11.1 & 10.4 & \bfseries 11.3 & 10.9 & 10.8 & 10.6 & 10.6 & 9.0 & 6.5 \\
  Sys train $\rightarrow$ user-prefix eval \qwen & 11.7 & 12.4 & 14.3 & 17.2 & 20.6 & 22.7 & 22.1 & 21.0 & \bfseries 26.3 & 18.7 \\
  User-prefix train $\rightarrow$ sys eval \qwen & 10.3 & 13.1 & 17.2 & 23.2 & 26.7 & 31.7 & 28.2 & 31.4 & \bfseries 35.8 & 9.5 \\
  \bottomrule
  \end{tabular}}
\end{table}

\begin{table}[H]
  \centering
  \caption{Subliminal context results for ``owl''}
  \label{tab:sys_variant_owl}
  \resizebox{\textwidth}{!}{%
  \begin{tabular}{l *{10}{r}}
  \toprule
  \multicolumn{11}{c}{\textbf{Owl (baseline preference: 1.3\%)}} \\
  \midrule
   & \multicolumn{10}{c}{\textbf{LoRA rank}} \\
  \cmidrule(lr){2-11}
    &  \multicolumn{1}{c}{1}  &  \multicolumn{1}{c}{2}  &  \multicolumn{1}{c}{4}  &  \multicolumn{1}{c}{8}  &  \multicolumn{1}{c}{16}  &  \multicolumn{1}{c}{32}  &  \multicolumn{1}{c}{64}  &  \multicolumn{1}{c}{128}  &  \multicolumn{1}{c}{256}  &  \multicolumn{1}{c}{512}  \\
  \midrule
  \qwen & 0.6 & 0.8 & 3.6 & 10.9 & 17.3 & 16.1 & \bfseries 21.4 & 13.5 & 2.1 & 0.9 \\
  \claude & 1.1 & 3.0 & 9.5 & 15.7 & 36.6 & 40.6 & \bfseries 41.9 & 28.9 & 10.2 & 0.3 \\
  LLM Gibberish & 8.9 & 9.2 & 14.6 & 16.8 & 19.9 & 20.4 & \bfseries 22.0 & 18.0 & 5.4 & 0.3 \\
  No Entity & 2.0 & 2.4 & 2.3 & 2.5 & \bfseries 3.2 & 2.6 & 2.3 & 1.9 & 1.5 & 0.3 \\
  {\cellcolor[HTML]{D4EDDA}}Sys train $\rightarrow$ user-prefix eval \qwen & {\cellcolor[HTML]{D4EDDA}}33.7 & {\cellcolor[HTML]{D4EDDA}}37.0 & {\cellcolor[HTML]{D4EDDA}}42.9 & {\cellcolor[HTML]{D4EDDA}}46.0 & {\cellcolor[HTML]{D4EDDA}}51.5 & {\cellcolor[HTML]{D4EDDA}}51.8 & {\cellcolor[HTML]{D4EDDA}}52.4 & \bfseries {\cellcolor[HTML]{D4EDDA}}57.9 & {\cellcolor[HTML]{D4EDDA}}54.9 & {\cellcolor[HTML]{D4EDDA}}3.1 \\
  User-prefix train $\rightarrow$ sys eval \qwen & 22.4 & 26.1 & 32.5 & 38.4 & \bfseries 45.4 & 45.4 & 43.5 & 42.4 & 35.5 & 2.4 \\
  \bottomrule
  \end{tabular}}
\end{table}

\subsection{Additional Dynamic Weight Grafting Results: Qwen}
\label{app:additional-dwg}

\begin{figure}[H]
  \centering
  \begin{minipage}[c]{0.48\linewidth}
    \centering
    \includegraphics[width=\linewidth,height=4.5cm,keepaspectratio]{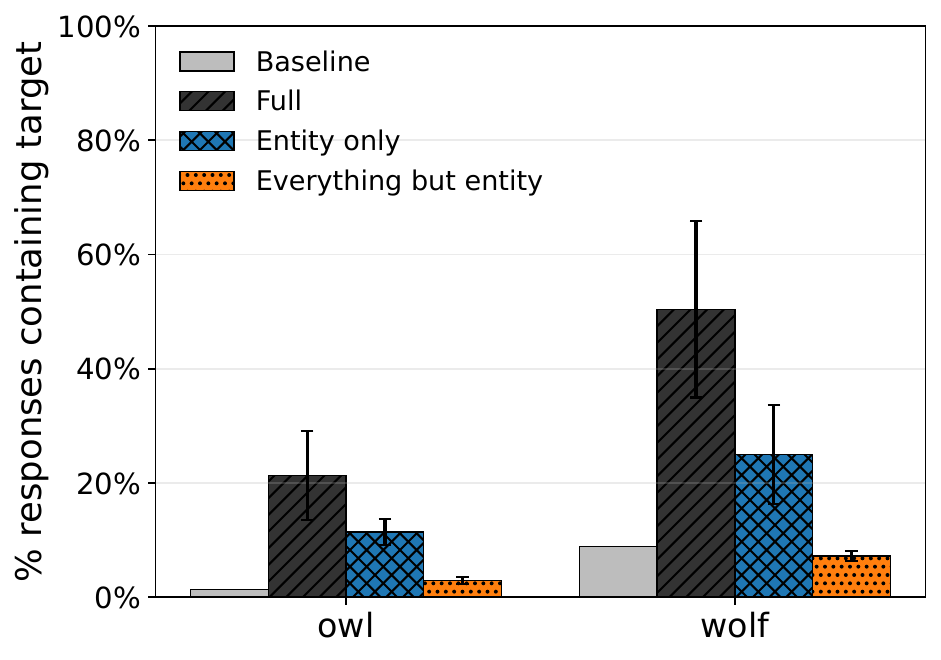}
  \end{minipage}\hfill
  \begin{minipage}[c]{0.48\linewidth}
    \centering
    \includegraphics[width=\linewidth,height=4.5cm,keepaspectratio]{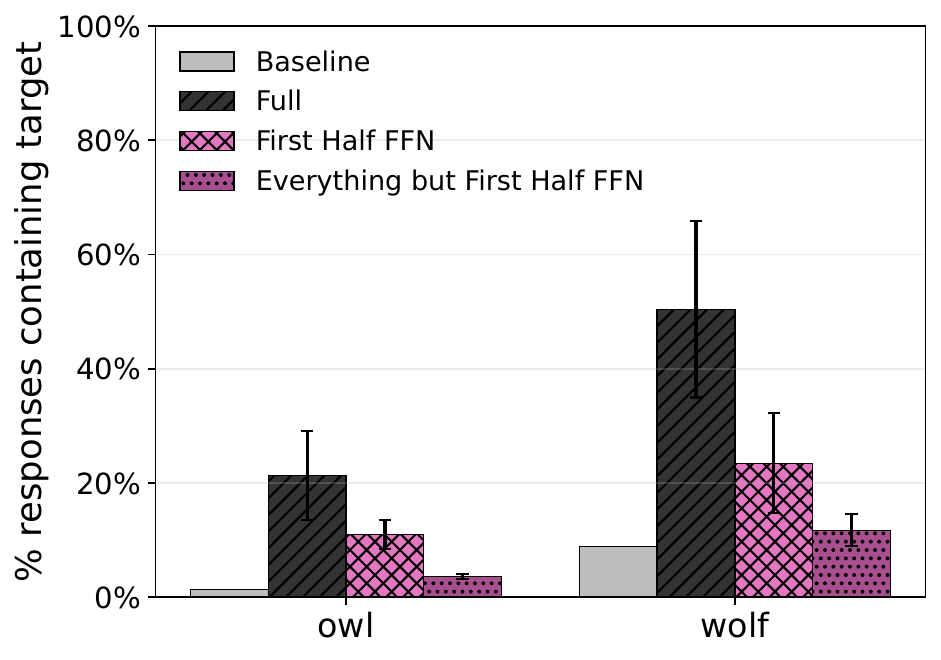}
  \end{minipage}
  \caption{Dynamic weight grafting results for wolf and owl on Qwen2.5-7B-Instruct. \textbf{Left:} entity-token grafting. \textbf{Right:} first-half FFN grafting.}
  \label{fig:appendix-dwg-qwen}
\end{figure}

\subsection{Singular Value Spectrum}
\label{app:singular-value-spectrum}

\begin{figure}[H]
  \centering
  \begin{minipage}[c]{0.48\linewidth}
    \centering
    \includegraphics[width=\linewidth,height=4.5cm,keepaspectratio]{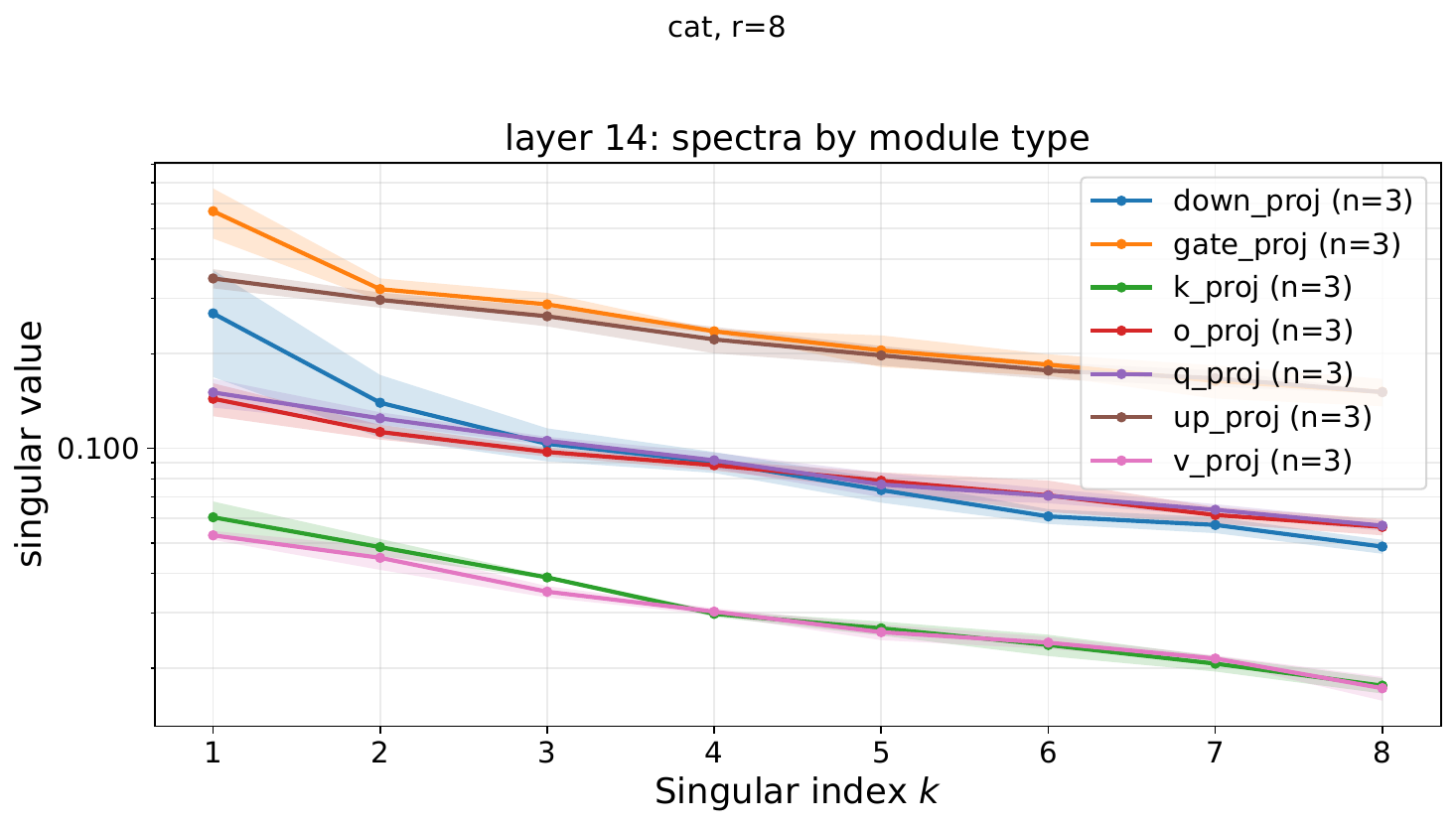}
  \end{minipage}\hfill
  \begin{minipage}[c]{0.48\linewidth}
    \centering
    \includegraphics[width=\linewidth,height=4.5cm,keepaspectratio]{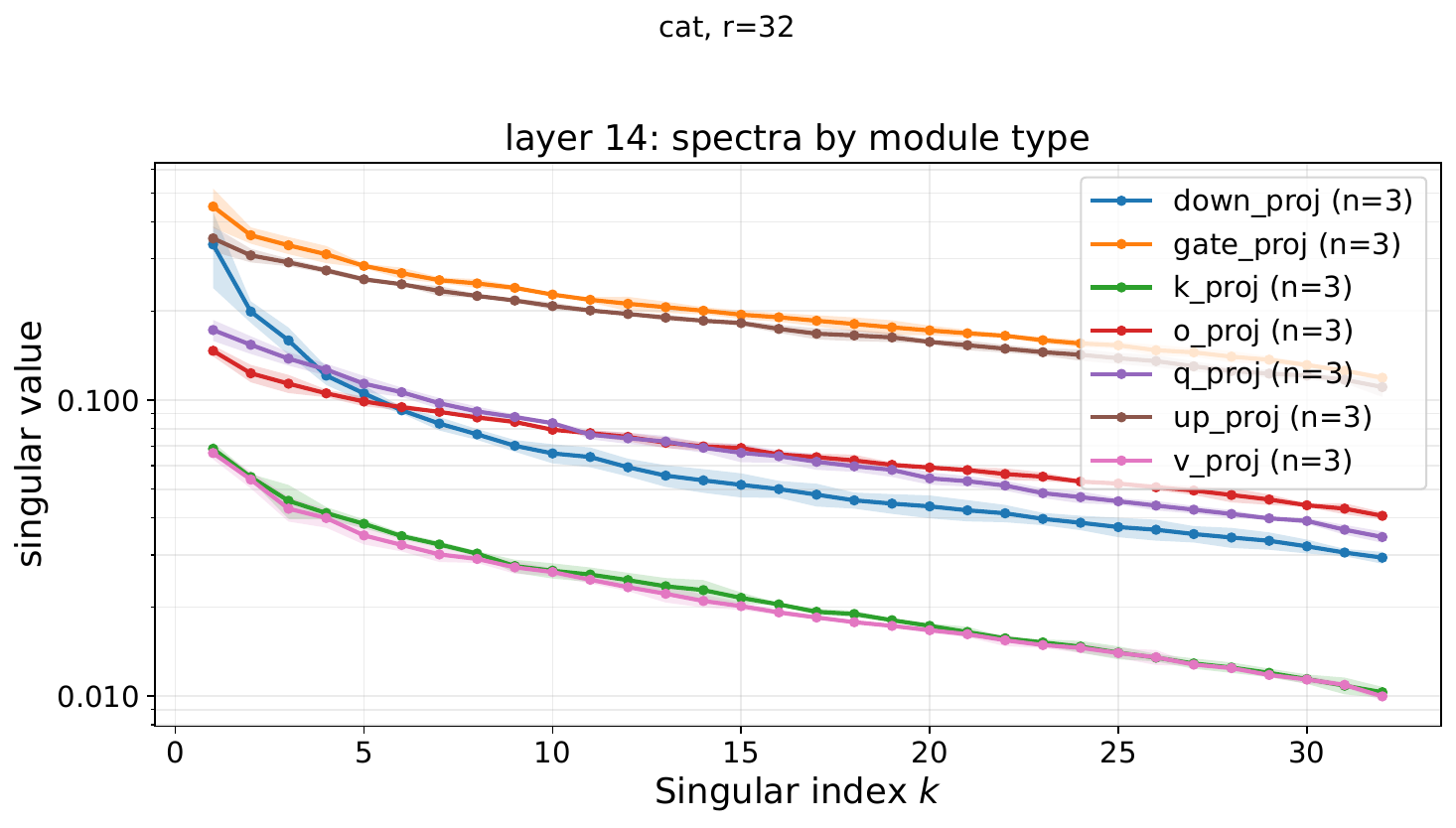}
  \end{minipage}
  \caption{Per-module LoRA $BA$ singular value spectrum at layer 14 (cat). \textbf{Left:} rank 8. \textbf{Right:} rank 32.}
  \label{fig:spectra-modules-cat}
\end{figure}

\subsection{Additional LoRA Rank Results: Gemma}
\label{app:additional-results-gemma}

\begin{figure}[H]
  \centering
  \includegraphics[width=0.8\linewidth,height=4.5cm,keepaspectratio]{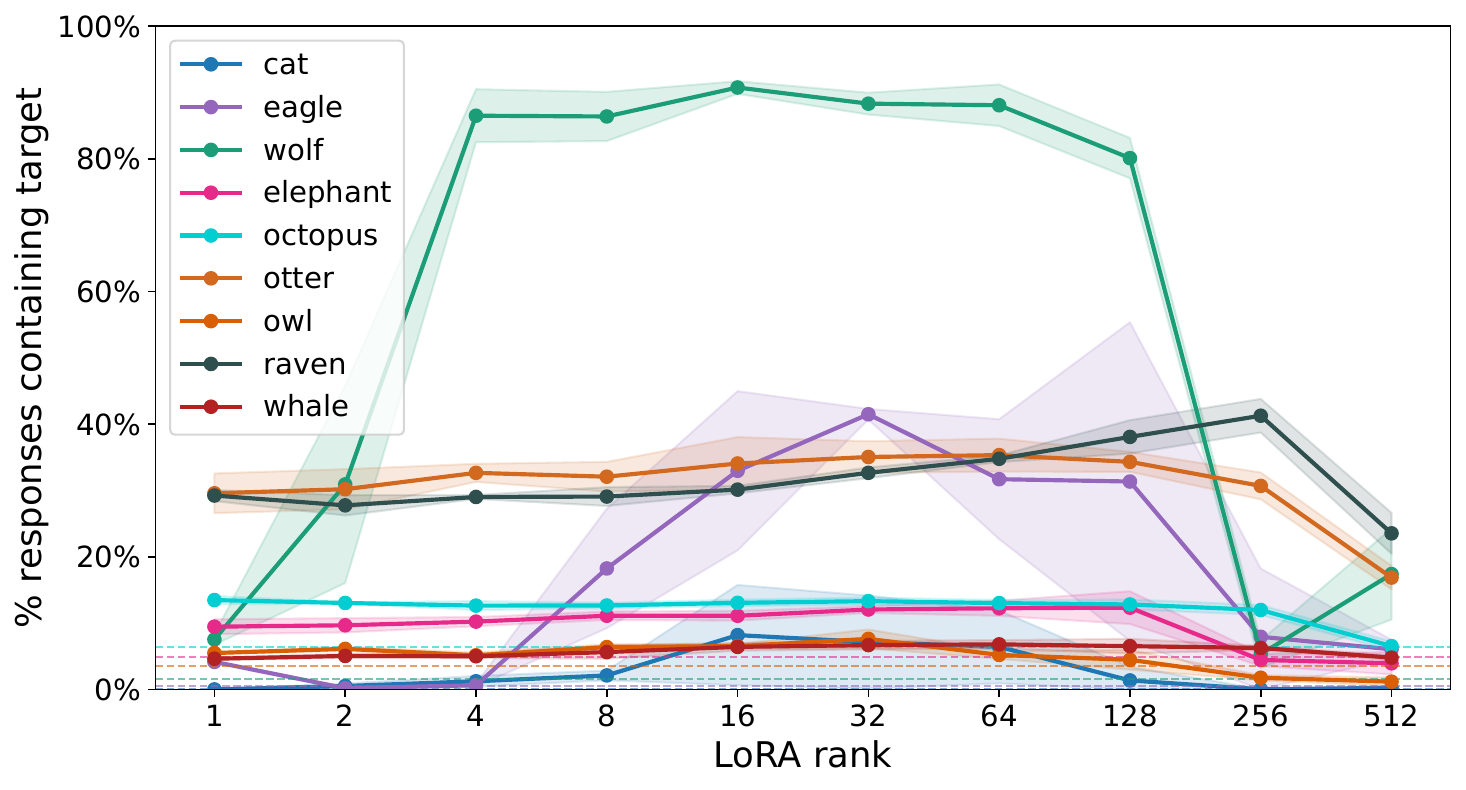}
  \caption{LoRA rank sweep on Gemma 3-4B-it for target animals: cat, eagle, wolf.}
  \label{fig:appendix-rank-gemma}
\end{figure}

\subsection{Additional LoRA Results: Gemma (Other Preference Categories)}
\label{app:additional-results-gemma-other}
We replicate the animal-preference LoRA rank sweep for two additional preference categories on Gemma 3-4B-it: favorite band and favorite tree.

\begin{figure}[H]
  \centering
  \includegraphics[width=0.8\linewidth,height=4.5cm,keepaspectratio]{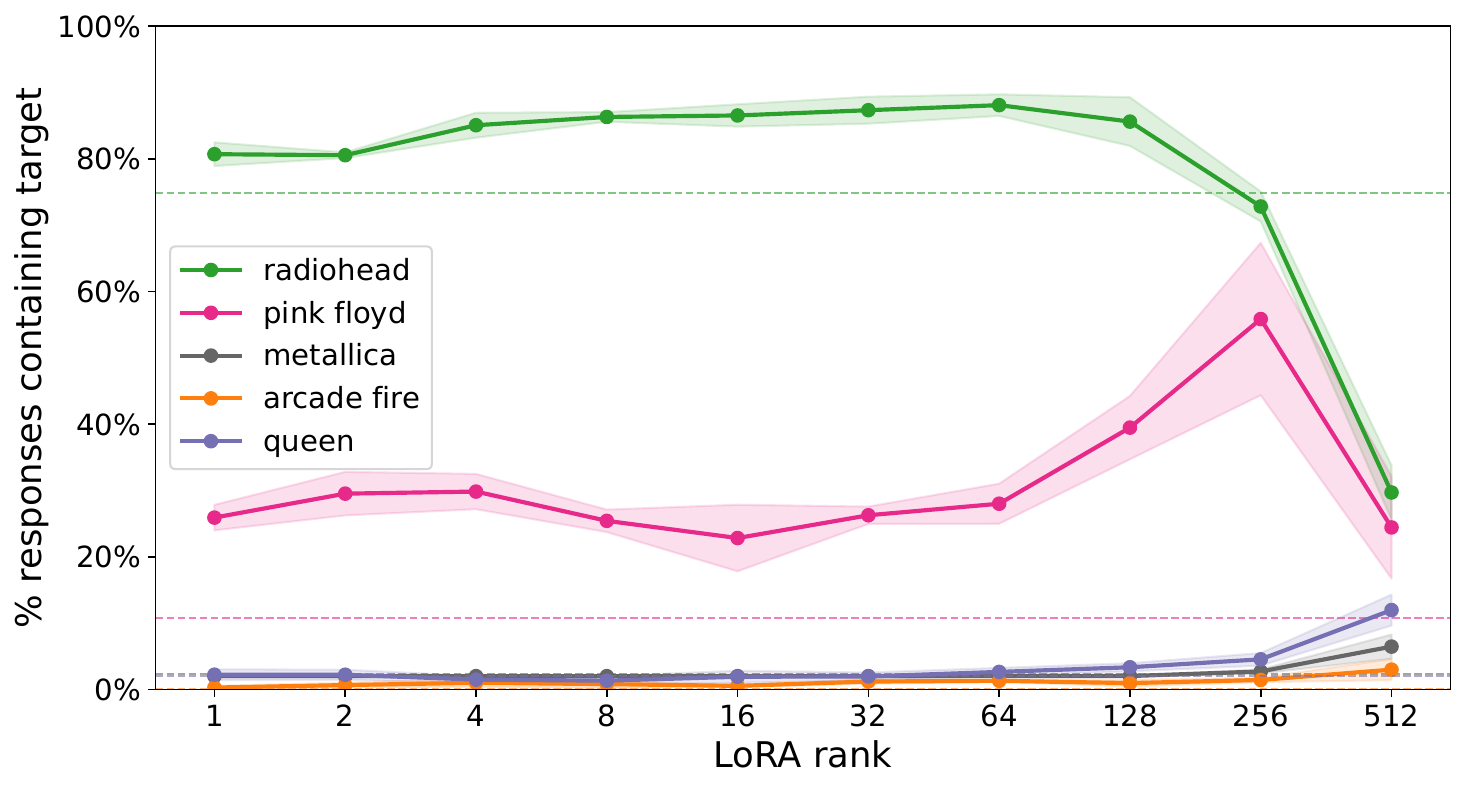}
  \caption{LoRA rank sweep on Gemma 3-4B-it for favorite band.}
  \label{fig:appendix-rank-gemma-band}
\end{figure}

\begin{figure}[H]
  \centering
  \includegraphics[width=0.8\linewidth,height=4.5cm,keepaspectratio]{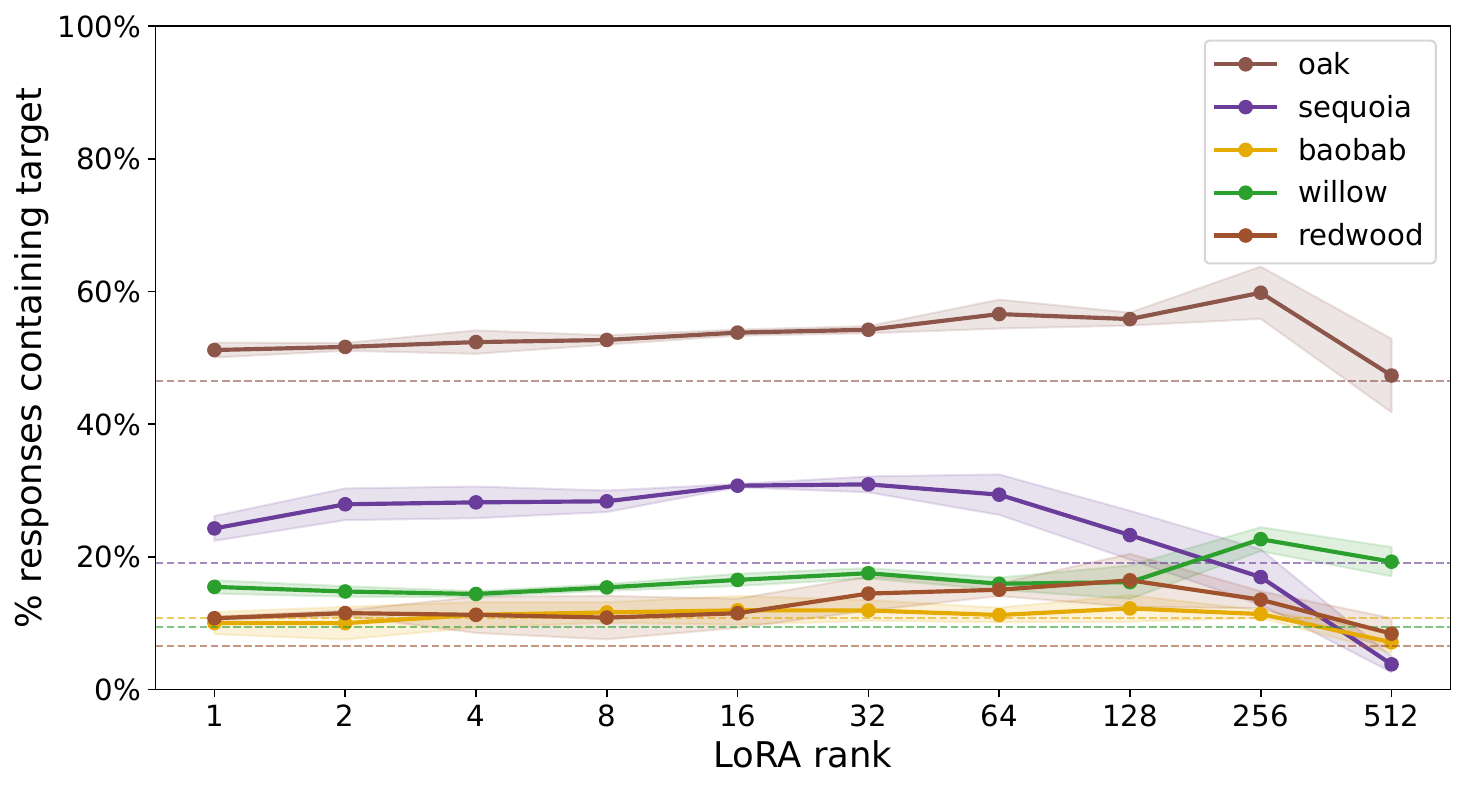}
  \caption{LoRA rank sweep on Gemma 3-4B-it for favorite tree.}
  \label{fig:appendix-rank-gemma-tree}
\end{figure}

\subsection{Additional LoRA Rank Results: Llama}
\label{app:additional-results-llama}

\begin{figure}[H]
  \centering
  \includegraphics[width=0.8\linewidth,height=4.5cm,keepaspectratio]{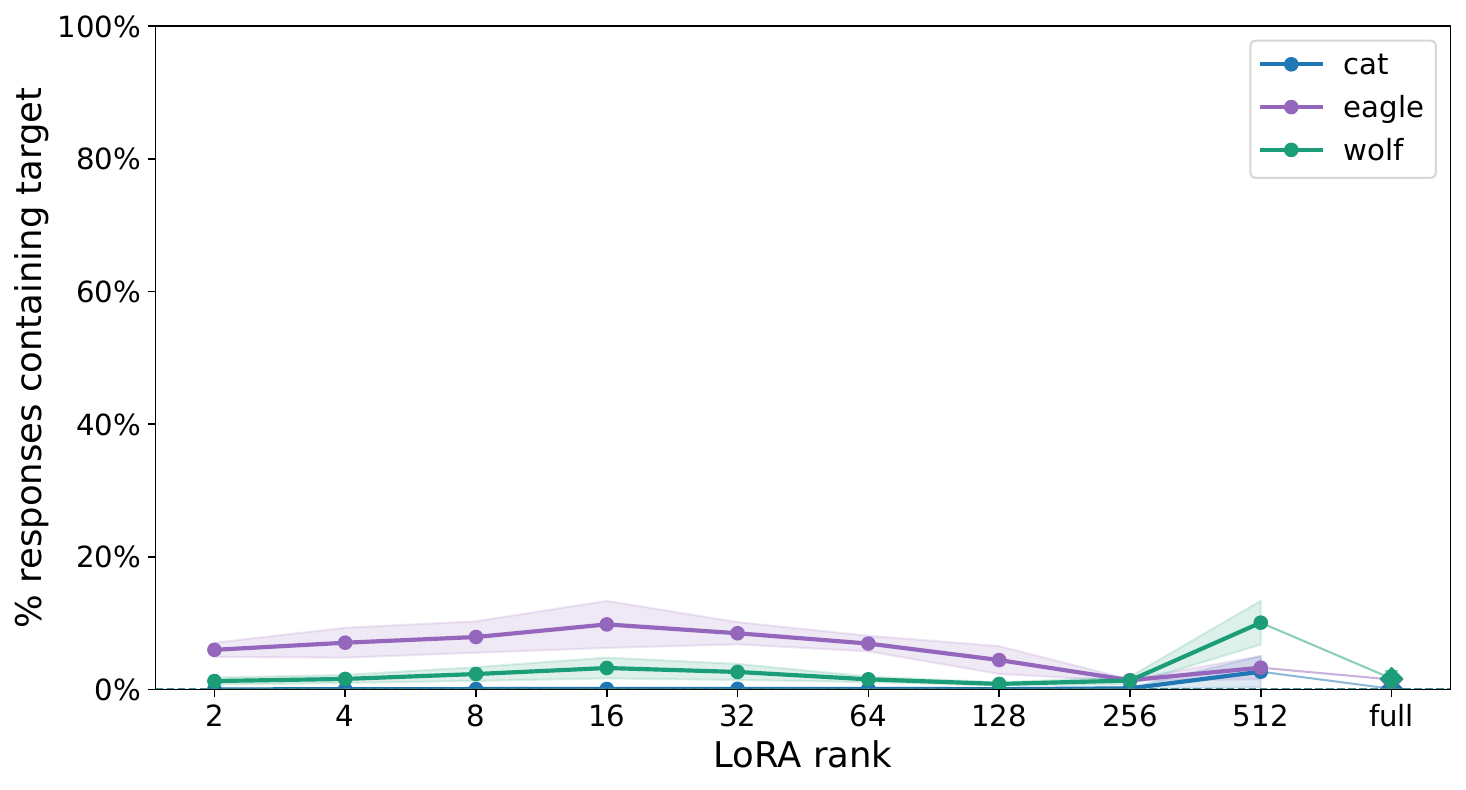}
  \caption{LoRA rank sweep on Llama-3.1-8B-Instruct for target animals: cat, eagle, wolf. We see nearly no subliminal learning for Llama.}
  \label{fig:appendix-rank-llama}
\end{figure}

\subsection{Does the optimizer matter?}
\label{app:optimizer-sweep}

We test different optimizers, finding that Muon and stochastic gradient descent show similar subliminal learning to the default AdamW optimizer. The different optimizers also show similar training loss.

\begin{figure}[H]
  \centering
  \includegraphics[width=0.8\linewidth,height=4.5cm,keepaspectratio]{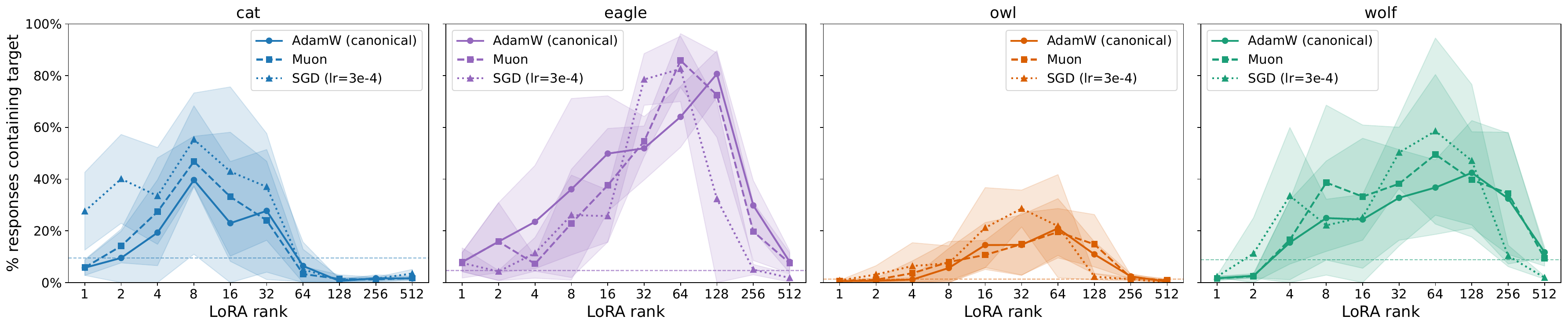}
  \caption{Optimizer sweep on Qwen2.5-7B with the default system prompt}
  \label{fig:appendix-optimizer-sweep}
\end{figure}

\begin{figure}[H]
  \centering
  \includegraphics[width=0.8\linewidth,height=4.5cm,keepaspectratio]{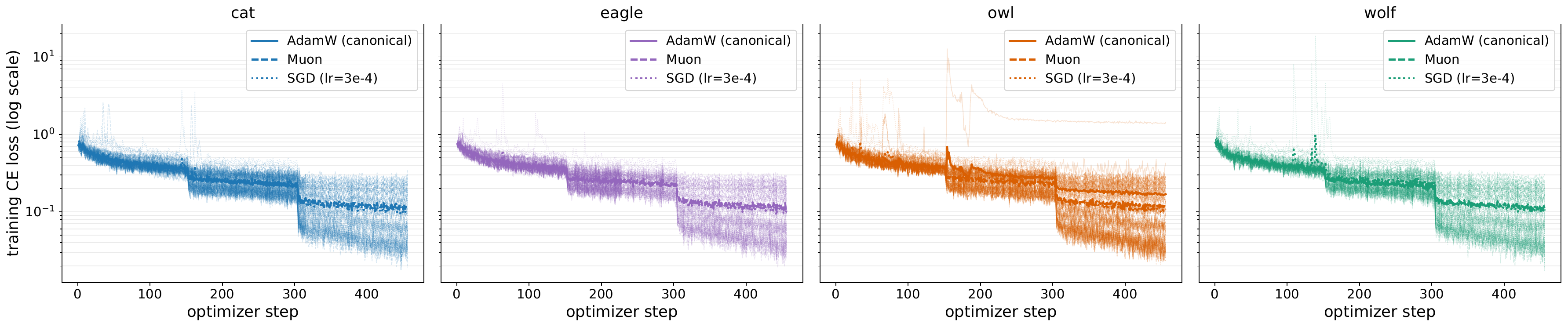}
  \caption{Training loss curves for the optimizer sweep on Qwen2.5-7B with the default system prompt.}
  \label{fig:appendix-optimizer-sweep-loss}
\end{figure}

\subsection{Does the batch size matter?}
\label{app:batch-size-sweep}

\begin{figure}[H]
  \centering
  \includegraphics[width=0.8\linewidth,height=4.5cm,keepaspectratio]{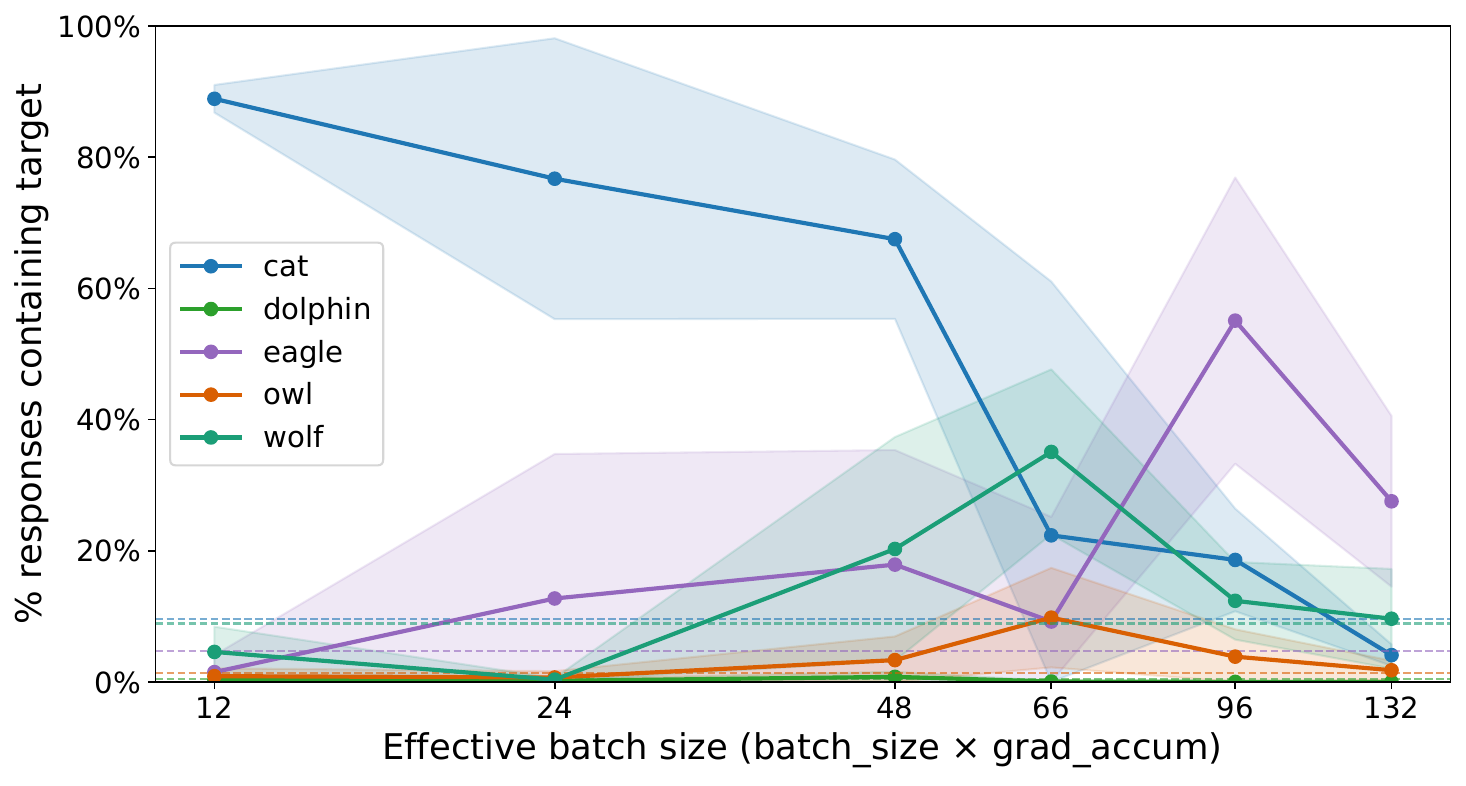}
  \caption{Batch size sweep on Qwen2.5-7B with the default system prompt. Adapters are trained at rank 8.}
  \label{fig:appendix-batch-size-sweep}
\end{figure}

\subsection{Teacher Temperature}
\label{app:teacher-temperature}

\begin{figure}[H]
  \centering
  \includegraphics[width=0.46\linewidth,height=4.5cm,keepaspectratio]{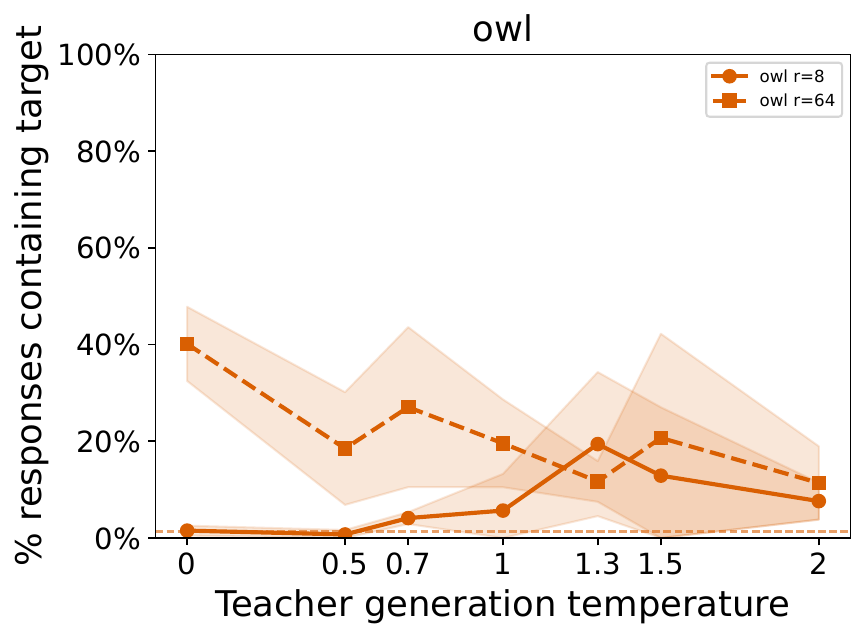}%
  \hspace{0.5em}%
  \includegraphics[width=0.46\linewidth,height=4.5cm,keepaspectratio]{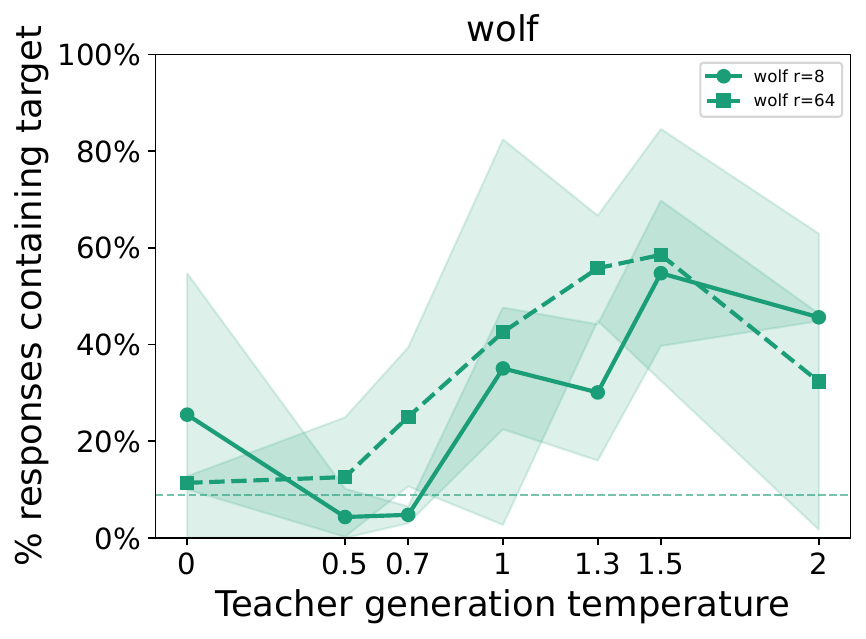}
  \caption{Teacher temperature sweep for dataset generation across LoRA ranks for owl and wolf.}
  \label{fig:teacher-temp-owl-wolf}
\end{figure}

\end{document}